\documentclass[12pt]{article}
\usepackage{amsmath,amsthm,amssymb}
\usepackage{mathtools}
\usepackage{mathrsfs}
\usepackage{times}
\usepackage{graphicx,subfigure}
\usepackage{color}
\usepackage{multirow}
\usepackage{apacite}
\usepackage[authoryear]{natbib}
\usepackage{rotating}
\usepackage{bbm}
\usepackage{latexsym}
\usepackage{epstopdf}
\usepackage{comment}
\usepackage{algorithm}
\usepackage{algorithmic}

\newtheorem{definition}{Definition}
\newtheorem{prop}{Proposition}

\newtheorem{cor}{Corollary}

\newcommand{\captionfonts}{\normalsize}

\makeatletter  
\long\def\@makecaption#1#2{%
  \vskip\abovecaptionskip
  \sbox\@tempboxa{{\captionfonts #1: #2}}%
  \ifdim \wd\@tempboxa >\hsize
    {\captionfonts #1: #2\par}
  \else
    \hbox to\hsize{\hfil\box\@tempboxa\hfil}%
  \fi
  \vskip\belowcaptionskip}
\makeatother   

\begin{document}
\hspace{13.9cm}1

\ \vspace{20mm}\\
{\LARGE Shapley Homology: Topological Analysis of Sample Influence for Neural Networks}

\ \\
{\bf Kaixuan Zhang$^{\displaystyle 1}$, 
Qinglong Wang$^{\displaystyle 2}$,
Xue Liu$^{\displaystyle 2},$
C. Lee Giles$^{\displaystyle 1}$}
\\
{$^{\displaystyle 1}$Pennsylvania State University.}\\
{$^{\displaystyle 2}$McGill University.}\\


\thispagestyle{empty}
\markboth{}{NC instructions}
\ \vspace{-0mm}\\
\begin{center} {\bf Abstract} \end{center}
Data samples collected for training machine learning models are typically assumed to be independent and identically distributed (iid).
Recent research has demonstrated that this assumption can be problematic as it simplifies the manifold of structured data. 
This has motivated different research areas such as data poisoning, model improvement, and explanation of machine learning models. 
In this work, we study the influence of a sample on determining the intrinsic topological features of its underlying manifold. We propose the Shapley Homology framework, which provides a quantitative metric for the influence of a sample of the homology of a simplicial complex. By interpreting the influence as a probability measure, we further define an entropy which reflects the complexity of the data manifold. Our empirical studies show that when using the $0$-dimensional homology, on neighboring graphs, samples with higher influence scores have more impact on the accuracy of neural networks for determining the graph connectivity and on several regular grammars whose higher entropy values imply more difficulty in being learned. 
 
\section{Introduction}
\label{intro}

In much machine learning research, it is common practice to assume that the training samples are independent and identically distributed (iid). 
As such, samples are implicitly regarded as having equal influence on determining the performance of a machine learning model. 
Recently, limitations of this assumption have been explored. 
For example, a recent study~\citep{KohL17Influence} showed that certain training samples can have significant influences over a model's decisions for certain testing samples. This effect has motivated research on model interpretation~\citep{gunning2017explainable,KohL17Influence,YehKYR18Representer}, model \& algorithm improvement~\citep{lee2019learning,Wang18Distillation,RenZYU18Reweight}, and poisoning attacks~\citep{Wang18Poisoning,KohL17Influence,Chen17Backdoor}.

Many of the aforementioned methods study the sample influence problem by using neural networks. 
Specifically, it is common to adopt a neural network (either a target model or a surrogate) to identify samples that are deemed as more influential~\citep{KohL17Influence,Wang18Distillation}. 
However, as shown by~\citet{Chen17Backdoor} in a poisoning attack scenario, a limited number of poisoning samples can be effective in applying a backdoor attack in a model-agnostic manner. 
This motivates work that studies the intrinsic properties of data so that one can develop countermeasures to methods that defeat learning models. 
Specifically, by representing the underlying space of data as a topological space, we study the sample influence problem from a topological perspective. 
The belief that topological features of the data space better represent its intrinsic properties has attracted recent research on establishing relationships between machine learning and topological data analysis (TDA)~\citep{Chazal17TDA,HoferKNU17,Carlsson18Topological}.

In this work, we consider TDA as a complementary approach to other approaches that study sample influence. 
We propose a framework for decomposing the topological features of a complex (built with data points and certain homology group) to individual points. 
We interpret the decomposed value for each point as representing its influence on affecting its topological features. 
We then calculate that influence as Shapley values~\citep{narahari2012shapley}. 
Recent research~\citep{Chen18LShapley,LundbergL17,DattaSZ16} proposed similar approaches to quantify feature influence. 
By interpreting the influence as a probability measure defined on a data space, we also calculate the entropy which describes the complexity of this space. 
Under our framework and with proper configurations of the topological space, we devise an algorithm for calculating the influence of data samples and entropy for any data set.



We perform both analytical and empirical studies on several sets of data samples from two families of structured data -- graphs and strings. 
Specifically, we generate random graphs using the Erdos-Renyi model and binary strings with regular grammars. 
In both cases, we use neural networks as verification tools for our analysis of sample influence. 
Explicitly, we employ a feed-forward network trained to determine the connectivity of the generated graphs, and a recurrent network trained to recognize the generated strings. 
Our results show that samples identified by our method as having more significant influence indeed have more impact on the connectivity of their underlying graphs, and that grammars with higher complexity have more difficulty in being learned. 



\section{Related Work}
\label{sec:rw}
\subsection{Sample Influence}
There have been several research efforts to explore and exploit influential samples for various purposes. 
As an example, by analyzing how the performance of a model at the testing phase affects a small number of training samples motivated research in poisoning attacks~\citep{Wang18Poisoning,KohL17Influence,Chen17Backdoor}. 
In this case, a limited number of corrupted training examples are injected to degrade a target model’s performance. 
An alternative thread of research exploits this effect to improve the generalization accuracy of learning models and the efficiency of learning algorithms. 
Specifically, influential samples can be identified via learning then used for enhancing models' performance on imbalance and corrupted data~\citep{RenZYU18Reweight,lee2019learning}. 
They can also be synthesized to represent a much larger set of samples hence accelerating the learning process~\citep{Wang18Distillation}. 
Besides, influential samples can bring explainability to deep learning models by identifying representative samples or prototypes used in decision making~\citep{YehKYR18Representer, anirudh2017influential}.

\subsection{Topological Data Analysis}

The most widely-used tool from TDA is persistent homology~\citep{Chazal17TDA}, which an algebraic method for measuring topological features of shapes and functions. 
Persistent homology provides a multi-scale analysis of underlying geometric structures represented by persistence barcode or diagram. 
Most of the previous research using persistence homology focused on the classification task of some certain manifold constructed by data points~\citep{carlsson2008local,li2014persistence,turner2014persistent}, and mapping the topological signatures to machine learning representations~\citep{Carlsson18Topological,bubenik2015statistical}. Unlike previous work, our proposed framework Shapley Homology emphasizes the individual influence of vertex on the topological space and generalizing the global features of the data set.



 
\section{Shapley Homology}
\label{sec:SH}
Here, we first introduce the relevant concepts that are necessary for our Shapley Homology framework. 
We then provide the definitions of sample influence and entropy.  

\subsection{$\breve{\text{C}}$ech Complex and Homology Group}
\label{sec:complex_homology}

\paragraph{$\breve{\text{C}}$ech Complex}
\label{sec:cech_complex}
In topological analysis, the study of a set of data points is typically based on simplicial complexes, which provide more abstract generalizations of neighboring graphs that describe the structural relationships between data points. 
Here, we are particularly interested in the $\breve{\text{C}}$ech complex, which is an abstract simplicial complex. More formally, the definition of the $\breve{\text{C}}$ech complex is as follows.

\begin{definition}[$\breve{\text{C}}$ech complex]
\label{def:cech}
The $\breve{\text{C}}$ech complex $C_r(X)$ is the intersection complex or nerve of the set of balls of radius $r$ centered at points in $X$.
\end{definition}


In particular, given a finite point cloud $X$ in a metric space and a radius $r>0$, the $\breve{\text{C}}$ech complex $C_r(X)$ can be constructed by first taking the points in $X$ as a vertex set of $C_r(X)$. 
Then for each set $\sigma \subset X$, let $X \in C_r(X)$ if the set of $r$-balls centered at points of $X$ has a nonempty intersection. 
Note, that the $\breve{\text{C}}$ech complex has a property showing that the simplicial complex constructed with a smaller $r$ is a subcomplex of that with a larger $r$.

\paragraph{Homology}
\label{sec:homology}
The topological features of data space are typically formalized and studied through~\emph{Homology}, which is a classical concept in algebraic topology~\citep{Chazal17TDA}. 
In the following, we briefly introduce homology to the extent that is necessary for understanding its role in our framework.
\begin{definition}[Homology Group]
A chain complex  $(\mathbf{A.},\mathbf{d.})$ is a sequence of abelian groups or modules connected by homomorphisms $d_n:A_n\rightarrow A_{n-1}$, such that the composition of any two consecutive maps is the zero map. The $k$-th homology group is the group of cycles modulo boundaries in degree $k$, that is,
\vskip -0.1in
\begin{equation}
H_k := \frac{\ker{d_n}}{\textup{im}\, d_{n+1}},
\end{equation}
\vskip 0.1in
where $\ker$ and $\textup{im}$ denotes the kernel and image of homomorphism, respectively.
\vskip 0.1in
\end{definition}

Generally speaking, the $k$-th homology is a quotient group that indicates the $k$-dimensional independent features of the space $\mathcal{X}$. Particularly, when $k = 0$, we have the following proposition~\citep{hatcher2005algebraic}:
\begin{prop}[]
For any space $\mathcal{X}$, $H_0(\mathcal{X})$ is a direct sum of abelian groups, one for each path-component of $\mathcal{X}$.
\label{prop:H0}
\end{prop}

Specially, when a complex takes the form of a neighboring graph, then the $0$-homology denote the subgraphs that are connected in this graph. 
As for the $1$-homology $H_1$, it represents the number of genus in the given manifold.

\subsection{Sample Influence and Entropy}
\label{sec:influence_entropy}
Given a point cloud represented by a $\breve{\text{C}}$ech complex,  we study the influence of each data point on the topological features of this complex. 
This influence of a data point can be further interpreted as the probability that a unit change of the topological feature is caused by this point. 
More formally, denote a data set containing $n$ samples by $X=\{x_1,\cdots,x_n\}$, we have the following definition of sample influence.
\begin{definition}[Sample Influence]
The influence of any sample (or subset) $\{x\} \subset X$ is a probability measure $\mu$.
\end{definition}
We can then define a canonical entropy of the set $X$ as follows.
\begin{definition}[Entropy]
\label{def:entropy}
Given a probability measure $\mu$, the entropy of a dataset $X$ is defined as $H(X)=-\sum_{i=1}^{n}\mu(x_i)\log{\mu(x_i)}$.
\end{definition}

\begin{figure*}[t]
\begin{center}
\includegraphics[scale=0.7]{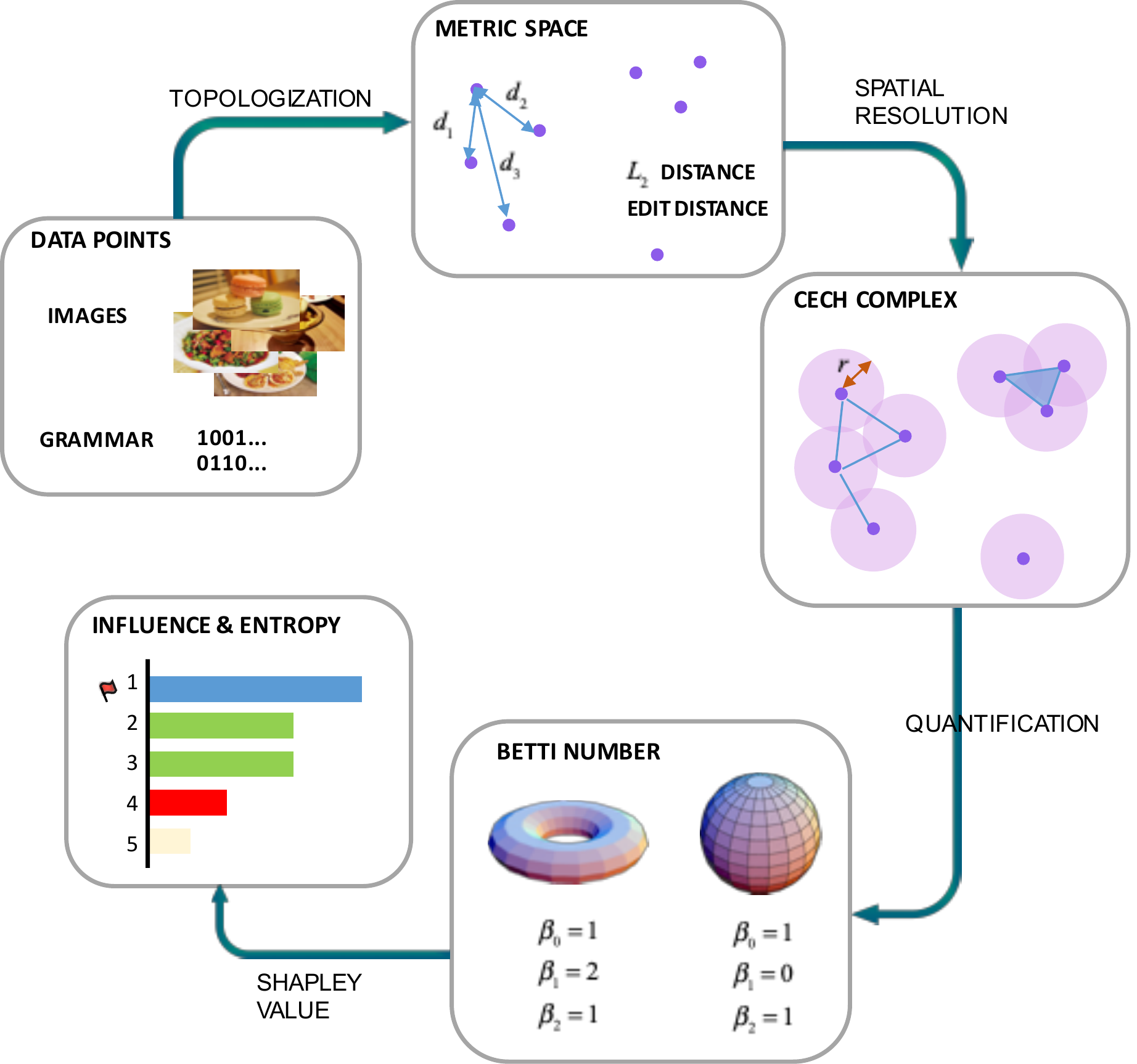}
\end{center}
\caption{Framework of Shapley Homology for influence analysis}
\label{fig:framework}
\end{figure*}

\subsection{The Framework of Shapley Homology}
\label{sec:framework}
Here we propose a framework (depicted in Figure~\ref{fig:framework}) of Shapley Homology in order to study sample influence. 
Specifically, in Figure~\ref{fig:framework}, we provide an example for investigating a specific topological feature -- the Betti number~\citep{rote2006computational} --of a topological space. 
As will become clear from the following, the Betti number can be used to quantify the homology group of topological space according to its connectivity.

\begin{definition}[Betti Number]
\label{def:betti}
Given a non-negative integer $k$, the $k$-th Betti number $\beta_k(\mathcal{X})$ of the space $\mathcal{X}$ is defined as the rank of the abelian group $H_k(\mathcal{X})$, that is,
\begin{equation}
\beta_k(\mathcal{X})=\textup{rank}(H_k(\mathcal{X})).
\end{equation}
\end{definition}

Following the definition, as there are no special structures (such as real projective space in the complex), the Betti number $\beta_k(\mathcal{X})$ indicates the number of direct sum of abelian groups of the $k$-th homology group\footnote{Generally speaking, the Betti number of a finitely generated abelian group $G$ is the (uniquely determined) number $n$ such that
\begin{equation}
\nonumber G=\mathbb{Z}^n\oplus G_1\oplus \cdots \oplus G_s,
\end{equation}
where $G_1, ..., G_s$ are finite cyclic groups. But for real projective space $\mathbb{RP}^n$, the homology group with coefficients in integers is
\begin{equation}
\setlength{\abovedisplayskip}{3pt}
\setlength{\belowdisplayskip}{3pt}
\nonumber H_k(\mathbb{RP}^n)=\left\{\begin{matrix}
\mathbb{Z} &  k=0\\ 
 \mathbb{Z}_2& k \textup{ odd},0<k<n\\ 
 0& \textup{otherwise}
\end{matrix}\right.
\end{equation}
Hence for odd $k$, the number of abelian groups in $H_k(\mathbb{RP}^n)$ is $1$ but actually the Betti number $\beta_k(\mathbb{RP}^n)=0$. 
This is not a concern for $\breve{\text{C}}$ech complex.}. 
In other words, the $k$-th Betti number refers to the number of $k$-dimensional holes on a topological surface. 

While the Betti number only indicates an overall feature of a complex, we need to further distribute this number to each vertex of this complex as its influence score. 
Recent research~\citep{Chen18LShapley} on interpreting neural networks has introduced Shapley value from cooperative game theory to decompose a classification score of a neural network made for a specific sample to individual features as their importance or contribution to rendering this classification result. 
Inspired by this line of work, we also employ the Shapley value as the influence score for each vertex. 
However, it should be noted that for a fixed $k$, the Betti number $\beta_k$ does not satisfy the monotonicity property of Shapley values. 
That is, the Betti number of $X_1$ is not necessarily larger than that of $X_2$ when $X_1$ is a subcomplex of $X_2$. As a result, we cannot adopt the formulation for calculating Shapley values directly. 
Here we use the following variant formulation for calculating Shapley values for a $\breve{\text{C}}$ech complex:
\begin{equation}
s(x_i)=\sum_{C \subseteq X\setminus x_i}\frac{\left | C \right |!(n-\left | C \right |-1)!}{n!}\left |\beta (C\cup x_i)-\beta(C))\right |.
\label{eq:shapley}
\end{equation}
It is important to note that in our formulation~\eqref{eq:shapley}, we use the absolute value to resolve the monotonicity issue. 
When the Betti number does satisfy the monotonicity property, then our calculation of Shapley values is the same as the canonical form. 

It is clear that our formulation~\eqref{eq:shapley} satisfies the symmetry axiom, whereas, it does not satisfy other Shapley axioms, including the linearity and carrier axioms~\citep{narahari2012shapley}. 
Nonetheless, our formulation still measures the marginal utility of a vertex. 
Besides, as we mainly focus on measuring the topological feature, both the decrease and increase in Betti number of a vertex are crucial for determining the influence. 
Furthermore, since our entropy is symmetry-invariant, its value will remain the same under the group action on the vertices~\citep{conrad2008group}.

Above discussion indicates that the influence of a data sample can be regarded as a function of the radius $r$, the Betti number $\beta_k$, the size of the data set containing this sample, and the topological space constructed upon the chosen metric. 
Unlike persistence homology~\citep{edelsbrunner2008persistent}, which studies the topological features of data space at varying ``resolution'' $r$, our analysis takes a more static view of the topological features of a complex built with a fixed $r$. 
As such, our analysis can be viewed as taking a slice from the filtration of used for persistence homology. 
In the following section, we propose an algorithm for calculating the influence scores of data points in a $\breve{\text{C}}$ech complex constructed with certain specified $r$ and $k$.

\section{Algorithm Design and Case Study}
\label{sec:algorithm_prepare}
With Proposition~\ref{prop:H0} and Definition~\ref{def:betti} and when the Betti number $\beta_0$ is regarded as a quantitative indicator, it equals the number of connected components in a complex. 
In this case, the task of calculating $\beta_0$ of a $\breve{\text{C}}$ech complex is equivalent to calculating the number of connected components of a graph. 
This enables us to compute the Laplacian matrix $L$~\footnote{The Laplacian matrix $L$ is defined to be $D-A$ where $D$ and $A$ denote the degree matrix and adjacency matrix respectively.} of a graph then apply the following proposition~\citep{marsden2013eigenvalues}.
\begin{prop}[]
A graph $G$ has $m$ connected components if and only if the algebraic multiplicity of \,$0$ in the Laplacian is $m$.
\label{prop:laplacian}
\end{prop}
With the above proposition, we can see that the Betti number $\beta_0(X)$ is equal to the number of zeros in the spectrum of the corresponding Laplacian matrix $L_X$. 
As such, we propose the Algorithm~\ref{algo:influence} to calculate the influence score and the entropy under the setting of $k = 0$ and $r$ as some constant. 
A more detailed discussion of the choice of $k$ and $r$ is provided in Section~\ref{sec:extended Discussion}. 
In the meantime, it is important to note that this Algorithm~\ref{algo:influence} always produces a probability measure. 

\begin{algorithm}[t]
\caption{}\label{algo:influence}
\begin{algorithmic}[1]

\REQUIRE ~~\\ 
Data $X=\{x_1,\dots,x_n\}$; Metric $d(\cdot,\cdot)$; Resolution $r$;
\ENSURE ~~\\ 
Influence $\mu(x_1),\dots,\mu(x_n)$; Entropy $H(D)$;\\

\FOR{each pair of data sample $x_i,x_j$} 
\STATE Calculate the pairwise distance $d(x_i,x_j)$;
\ENDFOR
\STATE Build the Complex $\Delta$  according to the resolution $r$;

\FOR{each subcomplex $C \subseteq \Delta$} 
\STATE Calculate the Laplacian matrix $L$ of $C$;
\STATE Calculate the multiplicity of eigenvalue 0 of of $L$;

\ENDFOR
\FOR{$i=1$ to $n$} 
\STATE Calculate $s(x_i)$ according to~\eqref{eq:shapley};
\ENDFOR
\STATE Set $\mu(x_i)$ as the normalized $s(x_i)$, then $H(X)=-\sum_{i=1}^{n}\mu(x_i)\log{\mu(x_i)}$;
\end{algorithmic}

\end{algorithm}

Here we provide a simple time complexity analysis for our algorithm and propose several possible solutions to accelerate the algorithm in our future work. 
Our algorithm can be decomposed into three parts: (1)~complex construction; (2)~calculation of graph spectrum; (3) assignment of the Shapley value. 
In the first step, since we need to calculate pairwise distance and form the adjacent matrix, it is clear that the complexity for this step is $O(n^2)$. 
In the second step, we need to compute the spectrum of all subcomplexes. 
As such, in total, the complexity is $O(n^3 2^n)$, where the complexity of each Laplacian decomposition is $O(n^3)$. 
As for the third step, we sum all the marginal utility for one sample, which results in the complexity of $O(2^n)$. 
Therefore, the complexity of computing the influence scores for all samples is $O(n 2^n)$. 
Based on the above analysis, we obtain the overall complexity of our algorithm as $O(n^3 2^n)$. 
Clearly, the second step and the third step contribute most to the total complexity. 
Specifically, in order to alleviate the computational burden caused by the second step, we will consider in our future work of various approaches (e.g.,~\citep{cohen2018approximating}) to approximate the spectrum of a graph. 
As for the third step, several existing approximation algorithms, e.g., C-Shapley and L-Shapley~\citep{Chen18LShapley}, could be considered for approximating the Shapley value using local topological properties.



In the following, we provide a set of case studies on several different graphs with representative topological structures. 
In particular, we study four types of graphs representing the space of binary strings generated by four regular grammars. 
We select these grammars due to their simplicity for demonstration. A brief introduction of the selected grammars~\footnote{Grammar 1, 2 and 4 are selected from the set of Tomita grammars~\citep{tomita1982dynamic}, of which their indices in the Tomita grammars are 1, 5 and 3, respectively.} is provided in Table~\ref{tab:grammars}. 
Since we deal with binary string, we specify the distance metric used in these studies as the edit distance~\citep{de2010grammatical} and set the radius $r = 1$. 
Also, we set the length $N$ of generated strings to fixed values as specified in Table~\ref{tab:grammars}. 

Furthermore, we generalize our proposed framework to another six special sets of graphs and provide the analytical results of their vertices' influence and the entropy values. These graphs are selected as they represent a set of simple complexes that can be used for building up more complicated topological structures. 

\begin{table}[t]
\small
\centering
\begin{tabular}{lll}
\hline \hline
$g$ & Description  & Entropy                              \\ \hline \hline
1 & $1^{*}$                                  & 0.00     \\ \hline
2 & even number of 0s and even number of 1s. & $\log{M}$     \\ \hline
3 & $1^{*} + 0^{*}(1+0)$                     & $3\log{2}/2$     \\ \hline
4 & \begin{tabular}[c]{@{}l@{}}an odd number of consecutive 1s is \\
    always followed by an even number\\of consecutive 0s.
    \end{tabular}                            & \begin{tabular}[c]{@{}l@{}}
    $\approx 2.30$\\($N=4$)   \end{tabular}  \\ \hline \hline
\end{tabular}
\caption{Example grammars and their associated entropy values.}
\label{tab:grammars}
\vskip 0.1in
\end{table}

\subsection{A Case Study on Regular Grammars}
\label{sec:study_grammar}
\begin{figure}[t]
\begin{center}
\includegraphics[scale=.6]{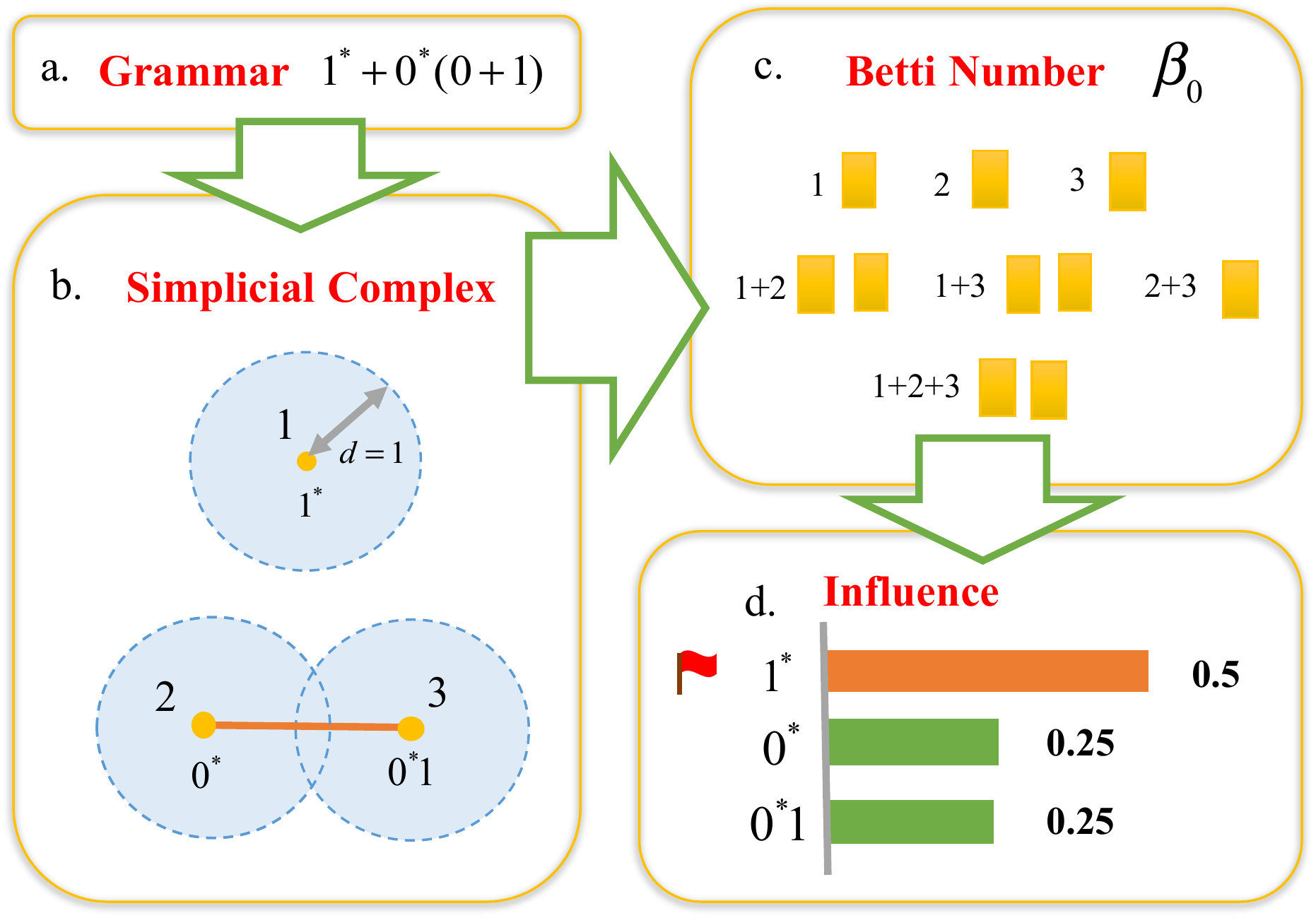}
\end{center}
\caption{An simple illustration of the Shapley Homology framework on grammar $1^{*}+0^{*}(0+1)$.}
\label{fig:example}
\end{figure}

\paragraph{Simple Examples} Here we calculate and analyze the first three grammars shown in Table~\ref{tab:grammars} as they have simple yet different topological features. 
As we can see from Table~\ref{tab:grammars}, for the first grammar, given a fixed $N$, there exists only one string defined by this grammar. 
As a result, the influence of this sample is $1$, and the entropy $H(g_1)=0$. 
For the second grammar, it is clear to see that any pair of valid strings have an edit distance larger than 1. 
In this case, the complex formed with the radius $r = 1$ consists of disjoint points. 
Assuming that the there exist $M$ valid strings defined by this grammar with the length of $N$, then all strings have the same influence $1/M$ and the entropy $H(g_2) = \log{M}$. 
For the third grammar, when the length $N$ of its associated strings is larger than 3, then the set $g_3$ of these strings can be expressed as  $g_3=\{1^N,0^N,0^{N-1}1\}$, denoted as $g_3=\{1,2,3\}$ for notation simplicity. 
We depict the complex for the case when $r=1$ in Figure~\ref{fig:example}. 
According to Proposition~\ref{prop:laplacian}, we then have the following Betti number $\beta_0$ of each subcomplex.
\begin{equation}
\begin{aligned}
\nonumber &\beta_0(G_1)=\beta_0(G_2)=\beta_0(G_3)=\beta_0(G_{2,3})=1,\\  &\beta_0(G_{1,2})=\beta_0(G_{1,3})=\beta_0(G_{1,2,3})=2.
 \end{aligned}
\end{equation}
where $G_S$ ($S \subseteq \{1,2,3\}$) denotes the subgraph of $G_{1,2,3}$ formed by the vertex set $S$. 
According to~\eqref{eq:shapley} and Algorithm~\ref{algo:influence}, we have  $\mu(1)=0.5, \mu(2)=\mu(3)=0.25$, and finally the entropy is $H(g_3)=\frac{3}{2}\log{2}$.

\begin{figure}[t]
\centering
  \subfigure[Resolution $r=1$.]{\label{fig:r1}\includegraphics[scale=0.55]{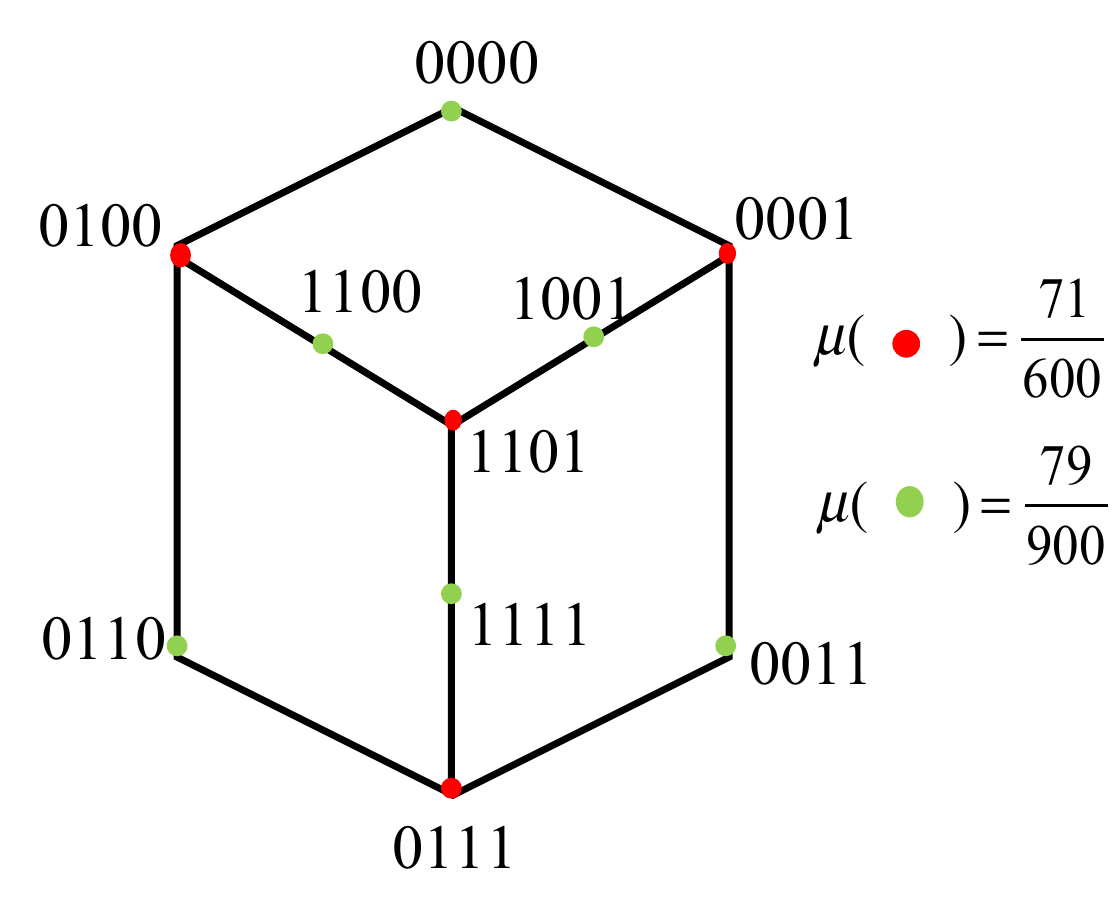}} \hfill
  \subfigure[Resolution $r=2$.]{\label{fig:r2}\includegraphics[scale=0.55]{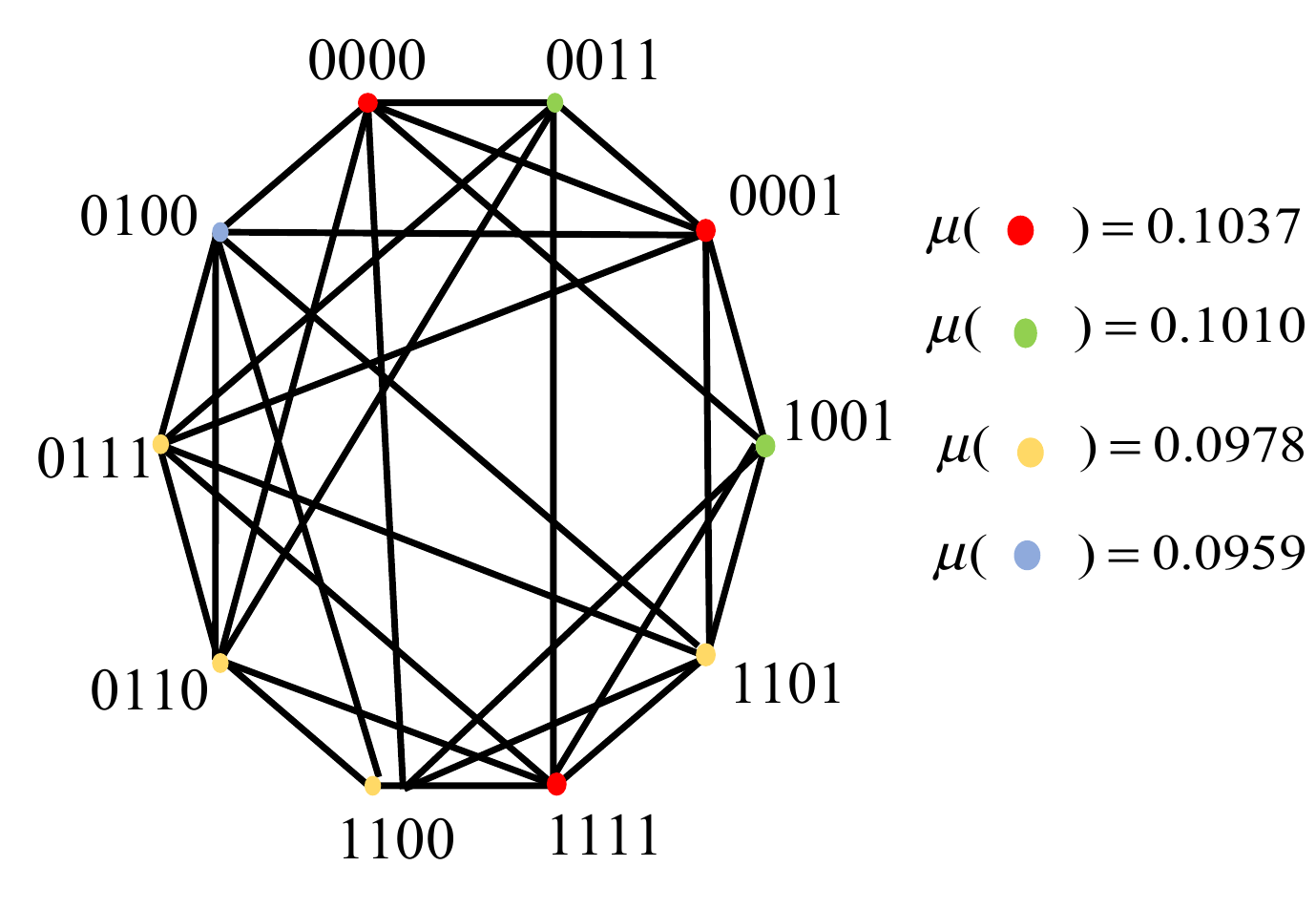}} \hfill \vskip -0.1in
\caption{The $\breve{\text{C}}$ech complex, influence scores and entropy of the fourth grammar with $N=4$.}
\label{fig:grammar3}
\vskip 0.1in
\end{figure}

\paragraph{A Complicated Example} 
The fourth grammar $g_4$ shown in Table~\ref{tab:grammars} is more complicated than the three grammars mentioned above. 
In particular, let $N=4$, $r=1$ or $2$, then we illustrate the results in Figure~\ref{fig:grammar3}, and the entropy is $H(g_4)=2.292$ when $r=1$ and $H(g_4)=2.302$ when $r=2$. 
Besides the analytical results presented here, we further demonstrate the difference between this grammar and the first two grammars in a set of empirical studies in Section~\ref{sec:exp_grammar}.

\begin{table*}
\scriptsize
\centering
\begin{tabular}{|c|c|l|l|}
\hline \hline
 & Example & Shapley Value & Influence Score \\ \hline \hline
\begin{tabular}[c]{@{}c@{}}Complete\\ Graph $K_n$\end{tabular} & \begin{tabular}[c]{@{}l@{}}\includegraphics[scale=0.15]{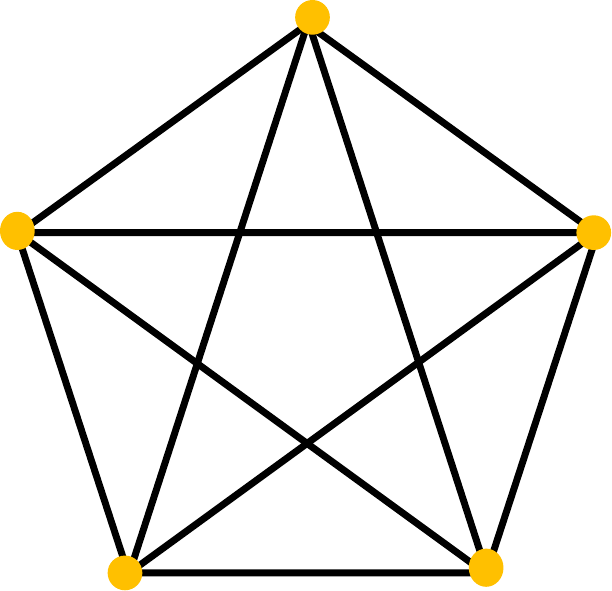} \end{tabular} & $1/n$ & $1/n$ \\ \hline
Cycles $C_n$ & \begin{tabular}[c]{@{}l@{}}\includegraphics[scale=0.15]{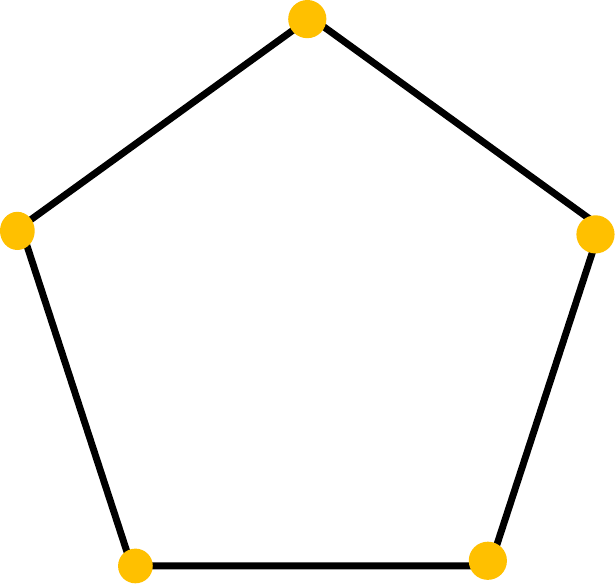}\end{tabular} & $2/3-1/n$ & $1/n$ \\ \hline
\multirow{2}{*}{\begin{tabular}[c]{@{}c@{}}Wheel\\ Graph $W_n$\end{tabular}} & \multirow{2}{*}{\includegraphics[scale=0.15]{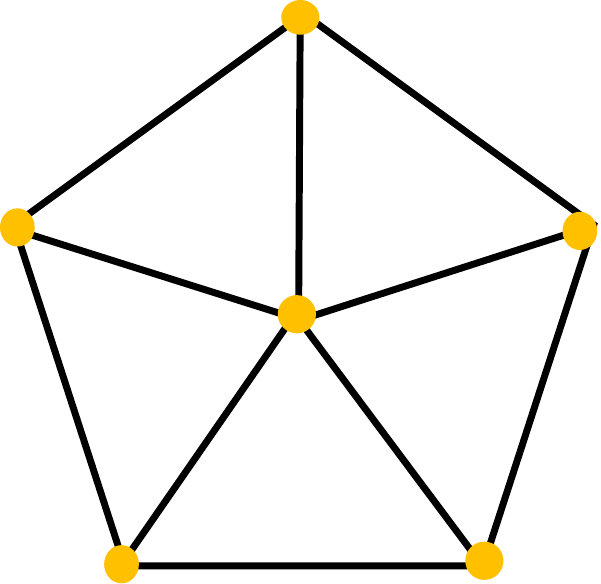}} & Periphery: $1/3 - 1/n(n-1)$ &  Periphery:  $\frac{2(n^2-n-3)}{3(n-1)(n^2-3n+4)}$  \\ \cline{3-4} 
 & & Center: $(n^2-7n+18)/6n$ & Center: $\frac{(n^2-7n+18)}{3(n^2-3n+4)}$  \\ \hline
\multirow{2}{*}{Star $S_{n-1}$} & \multirow{2}{*}{ \includegraphics[scale=0.15]{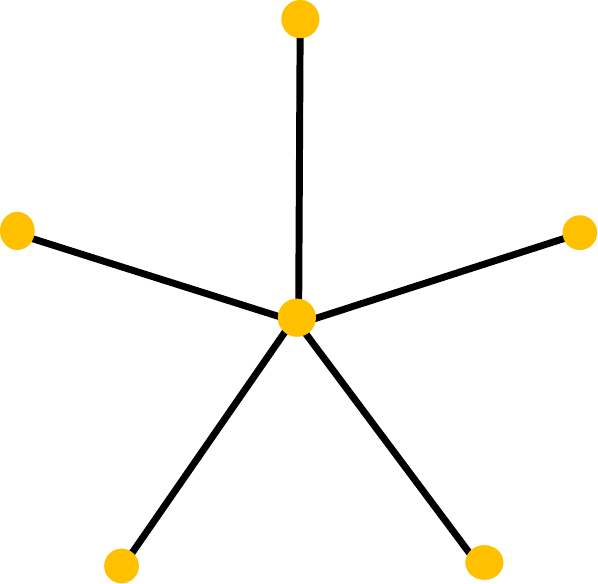}} &  Periphery: $1/2$           &  Periphery: $n/2(n^2-2n+2)$\\ \cline{3-4} 
 & & Center: $(n^2-3n+4)/2n$ & Center: $\frac{n^2-3n+4}{2(n^2-2n+2)}$   \\ \hline
\multirow{2}{*}{Path Graph $P_n$} &  \multirow{2}{*}{\includegraphics[scale=0.2]{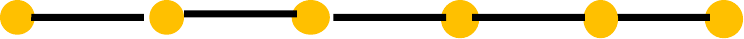} } & Ends: $1/2$ &  Ends:  $3/2(2n-1)$\\ \cline{3-4} 
 & & Middle: $2/3$ &  Middle: $2/(2n-1)$   \\ \hline
\multirow{2}{*}{\begin{tabular}[c]{@{}c@{}}Complete Bipartite\\Graph $K_{m,n}$\end{tabular}}   & \multirow{2}{*}{\includegraphics[scale=0.2]{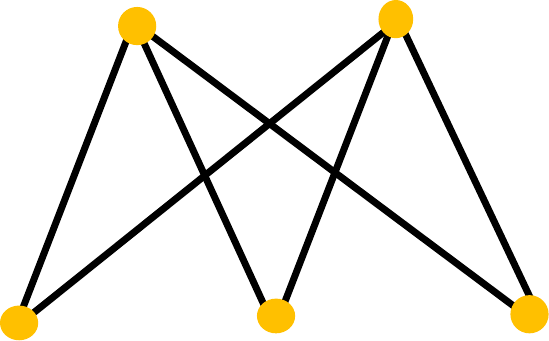} }    &  $m$ side:  $\frac{n(n-1)}{m(m+1)(m+n)}+\frac{1}{n+1}$  &    $m$ side: $\frac{m^3+n^3+m^2n+mn+m^2-n}{m(m+n)(2m^2+2n^2+m+n-mn-1)}$\\ \cline{3-4} 
                                                &      & $n$ side: $\frac{m(m-1)}{n(n+1)(m+n)}+\frac{1}{m+1}$     &   $n$ side: $\frac{m^3+n^3+n^2m+mn+n^2-m}{n(m+n)(2m^2+2n^2+m+n-mn-1)}$   \\ \hline \hline
\end{tabular}
\caption{Six special sets of graphs and their values of influence for the vertices.}
\label{tab:special graph results}
\vskip 0.1in
\end{table*}

\subsection{A Case Study on Special Graphs}
\label{sec:study_graph}
In this part, we apply our framework to six special families of graphs, which are shown with examples in Table~\ref{tab:special graph results}. Due to space constraint, here we omit the detailed calculation of the influence scores and entropy values for these graphs. With the analytical results shown in Table~\ref{tab:special graph results}, it is easy to derive the following two corollaries:
\begin{cor}[]
\label{cor1}
$H(K_n)=H(C_n)>H(W_n)>H(S_{n-1})$ for $n>5$. 
\end{cor}
\begin{cor}[]
\label{cor2}
Suppose that $m+n$ is constant, then the value of $H(K_{m,n})$ decreases as $\left |  m-n \right |$ increases.
\end{cor}
 
\section{Extended Discussion of Shapley Homology}
\label{sec:extended Discussion}
As previously introduced in Section~\ref{sec:framework}, our proposed influence score can be viewed as a function of the parameter $k$ and $r$. 
Here we extend our study from the previous case of $k = 0$ to cover other cases when $k > 0$, and show that adopting the $0$-dimensional homology offers the merits of having more meaningful and practical implication. 
Also, we discuss the effect of selecting different radii $r$ on the topological features we study. 
More importantly, we show that our framework provides a more general view of sample influence and can be further adapted to capture the dynamical features of a topological space. 

\subsection{Discussion of the Choice of $k$}
We first consider the case when $k = 1$, since the homology group becomes less informative as $k$ increases. 
In this case, the Betti number $\beta_1$ represents the number of genus in a given manifold. 
Take the third grammar introduced in Section~\ref{sec:study_grammar} as an example. 
It is clear to see that when using the $1$-dimensional homology, all subcomplex has $\beta_1$ equal to 0. 
This is precisely the case when our algorithm cannot generate a probability measure. 
In a more general case, it is easy to show by the homology theory~\citep{schapira2001categories} that once all subcomplex gets Betti number $\beta_{\hat{K}}=0$ for some $\hat{K}$, then for any other $k$-dimensional homology with $k > \hat{K}$, our framework cannot produce a probability measure. 
Furthermore, it is important to note that when adopting $k$-dimensional homology with $k$ larger than 2, the corresponding Betti number $\beta_k$ only has rather abstract interpretation.

Another difficulty of adopting homology with higher dimension is due to the practical concern. 
More specifically, since it is challenging to calculate the homology when $k$ is large, one may need to apply tools such as Euler characteristic or Mayer-Vietoris Sequences~\citep{hatcher2005algebraic}, which is out the scope of this paper.

\subsection{Discussion on the Choice of $r$}
The radius $r$ plays a critical role in building a manifold. 
For example, when $r$ is sufficiently small, the resulting complex contains only discrete data points with equal influence in our framework. 
When $r$ is sufficiently large, on the other hand, the complex is contractible, which indicates that there is no difference between the influence of data points contained in this complex. 
Both aforementioned extreme cases produce equal importance to the data points in a data set, and correspond to the i.i.d. assumption of sample distribution that is commonly taken in nowadays machine learning practice. 
In this sense, our framework provides a general abstraction of sample influence.

In practice, selecting a proper $r$ can be very difficult. 
This explains the usage of persistence homology~\citep{edelsbrunner2008persistent}, which studies the evolution of topological features of a space as $r$ dynamically changes. 
This motivates future research of extending our framework to dynamically generate a series of probability measures for the influence and the corresponding entropy values for a data set. 
However, we need to calculate the Shapely value instead of the persistence diagram during the process, and it is well known that calculating the exact Shapley value leads to computational difficulties. 
This sheds light on future research of exploiting the local topological structure of data points to accelerate the approximation process. 
Similar approaches have been proposed in decomposing a learning model's prediction result into feature importance~\citep{Chen18LShapley}.

\subsection{Remarks on Related Definitions of Entropy}
Different definitions of entropy have been proposed in previous research. Here we briefly revisit several representative definitions and compare with ours. 

\paragraph{Graph Entropy} 
An important property of the graph entropy~\citep{rezaei2013entropy} is monotonicity since it is defined based on mutual information. 
Specifically, it describes that the entropy of a subgraph is smaller than that of the whole graph on the same vertex set. 
In our case, by considering the complex as a graph, our entropy is defined to capture the geometric properties (such as the symmetry invariant property mentioned in Section~\ref{sec:framework}) of a graph. 
More specifically, our entropy measure focuses more on the variation of the topological features when a graph is changed. 
As such, our definition also covers variations that may violate the monotonicity property and the subadditivity property.

\paragraph{Entropy for Grammatical Learning} 
This entropy~\citep{wang2018comparative} is defined to reflect the balance between the population of strings accepted and rejected by a particular grammar. 
It shows that the entropy of a certain grammar is equal to that of the complement of that grammar. 
This raises a contradiction in the intuition that a grammar with a high cardinality of strings is more likely to have a higher entropy value. 
Our entropy, on the other hand, is defined to capture the intrinsic properties of a set of samples instead of reflecting the difference between different sets of samples. In this sense, our entropy is more like to assign a higher entropy value to a set of samples with larger cardinality.

\paragraph{Entropy in Symbolic Dynamics}
This type of entropy~\citep{williams2004introduction} is defined to reflect the cardinality of a shift space, which can be regarded as a more general definition of regular grammar. 
It implicitly assumes that any shift contained in a shift space has equal influence. 
This is contrary to our case in that we define the entropy to describe the complexity of a topological space that contains vertices with different influence. 
As such, our entropy provides a more fine-grained description of a topological space.

\subsection{Connection to the VC dimension}
Here we provide a preliminary discussion on the connection between our proposed Shapley Homology and Vapnik–Chervonenkis (VC) dimension~\citep{VCdimension}, which essentially reflects the complexity of a space of functions by measuring the cardinality of the largest set of samples that can be ``shattered'' by functions in this space. 
From a similar point of view, we expect the topology of data space is also critical in evaluating the complexity of real-world learning algorithms in learning a given set of data. 

Note that the proposed approach has a close connection to statistical learning theory. 
An obvious interpretation of our introduced complexity can be taken analogously as sample complexity~\citep{hanneke2016optimal}, which is closely related to the VC dimension. 
However, here we specify the limitation of the complexity specified by the VC dimension, which is part of the motivation of this work. Specifically, given a certain data space, only the hypothesis space of models with sufficiently large VC dimension can shatter this data space. 
For this hypothesis space, we can further use sample complexity to specify the number of samples required to achieve certain PAC learning criteria. 
However, we argue that different shattering of the data space leads to different levels of complexity (or different entropy values in our terms). 
Instead of focusing only on the maximally shattered data space, we argue that in practice, when building a learning model, a different shattering should be treated differently. 
To better explain this effect, we take regular grammars as an example case. 
One can consider a certain binary regular grammar as a certain configuration in the hypothesis space. 
In other words, a binary regular grammar explicitly split all the $2^N$ strings into the set of accept strings and the set of rejected strings, given that the strings have a fixed length of N. 
Since this grammar is equivalent to a DFA, and if we regard this DFA as a classification model, it itself has a certain VC dimension~\citep{ishigami1997vc}. 
Indeed, this effect is shown in the experiments in Section~\ref{sec:exp} on grammar learning. 
In particular, in our experiments, the demonstrated different levels of difficulty of learning different regular grammars indicate that different grammars (or different ways of shattering the data space) should not be taken as equal.

\section{Experiments}
\label{sec:exp}
In this section, we first demonstrate the results of using our Algorithm~\ref{algo:influence} to identify influential nodes in random graphs. 
Then we evaluate on several data sets generated by regular grammars to determine if data sets assigned by our algorithm to higher entropy values cause more challenges for neural networks to learn these grammars. 
The settings of all parameters in the experiments are provided in the supplementary file.

\subsection{Graph Classification}
\label{sec:exp_graph}
For these experiments, we first constructed the learning models and data sets of random graphs following \citet{dai2018adversarial}. 
We adopted this similar setting for the different purpose of evaluating and verifying the influence of individual nodes in a graph. 
To avoid the evaluation results from being biased by a particular synthetic data set, we performed 20 experimental trials by varying the probability used in the Erdos-Renyi random graph model to generate the same amount of random graphs. 
More specifically, in each trial, we generated a set of 6,000 undirected random graphs from three classes (2000 for each class), and split the generated set into a training and a testing set with the ratio of 9 to 1. 

\begin{figure}[t]
\centering
  \subfigure[The original graph]{\label{fig:g0}\includegraphics[width=40mm]{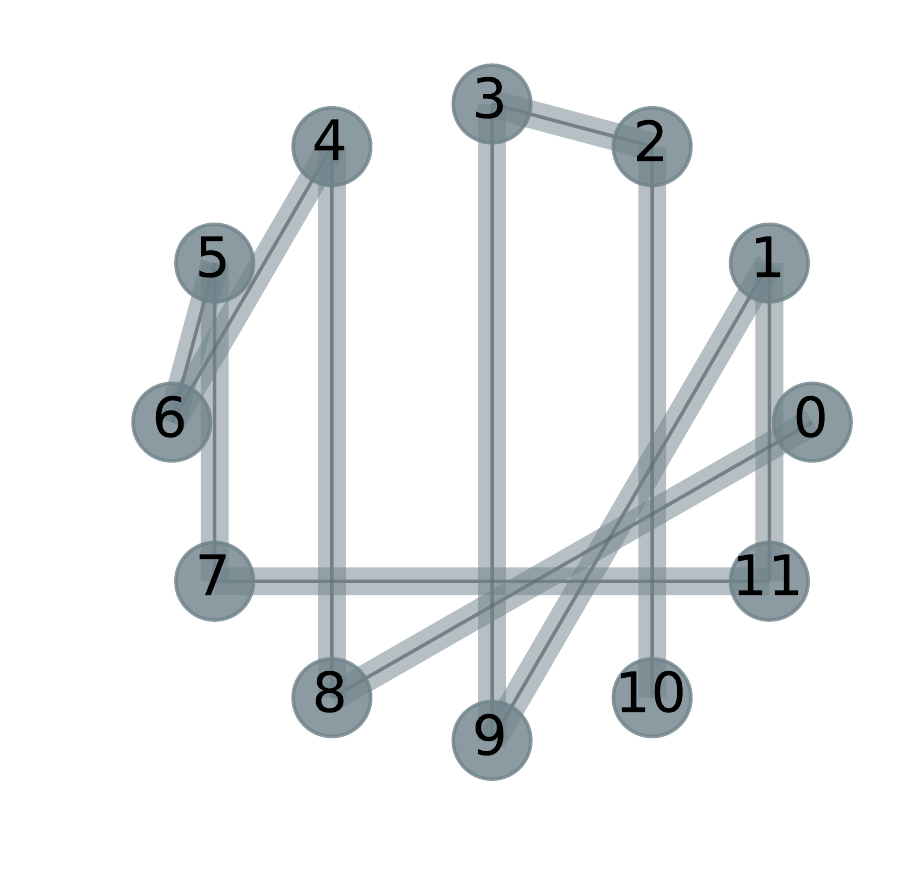}} \hfill 
  \subfigure[Graph manipulated by masking the most influential node.]{\label{fig:gmax}\includegraphics[width=40mm]{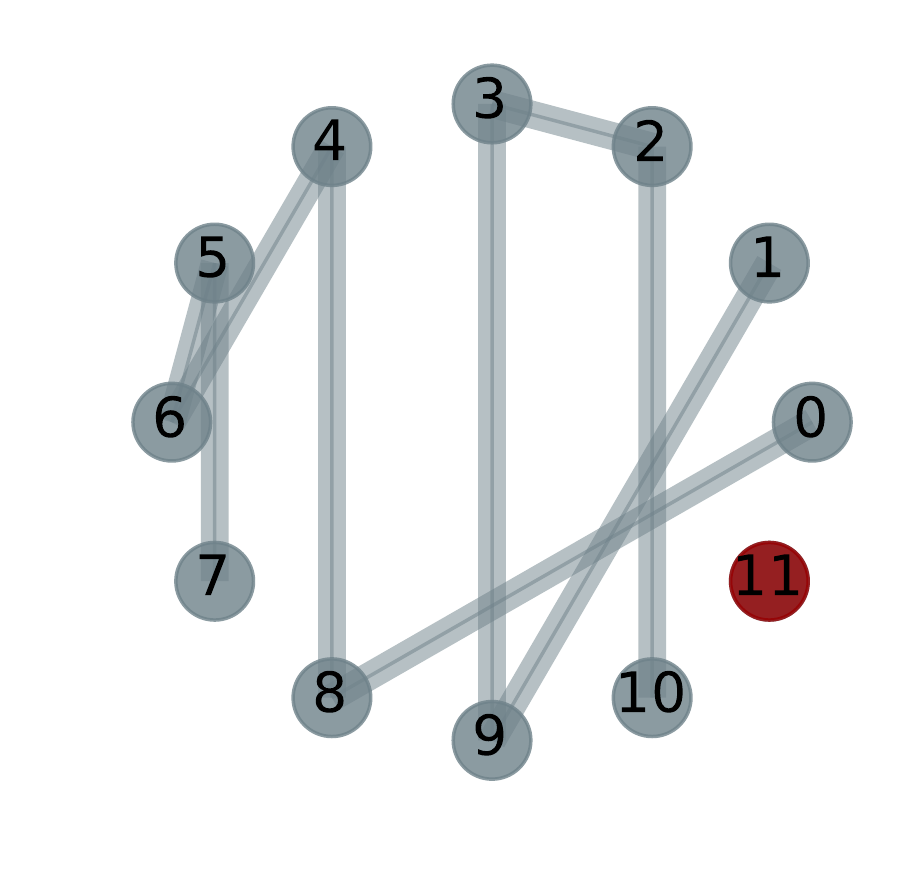}} \hfill 
  \subfigure[Graph manipulated by masking the least influential node.]{\label{fig:gmin}\includegraphics[width=40mm]{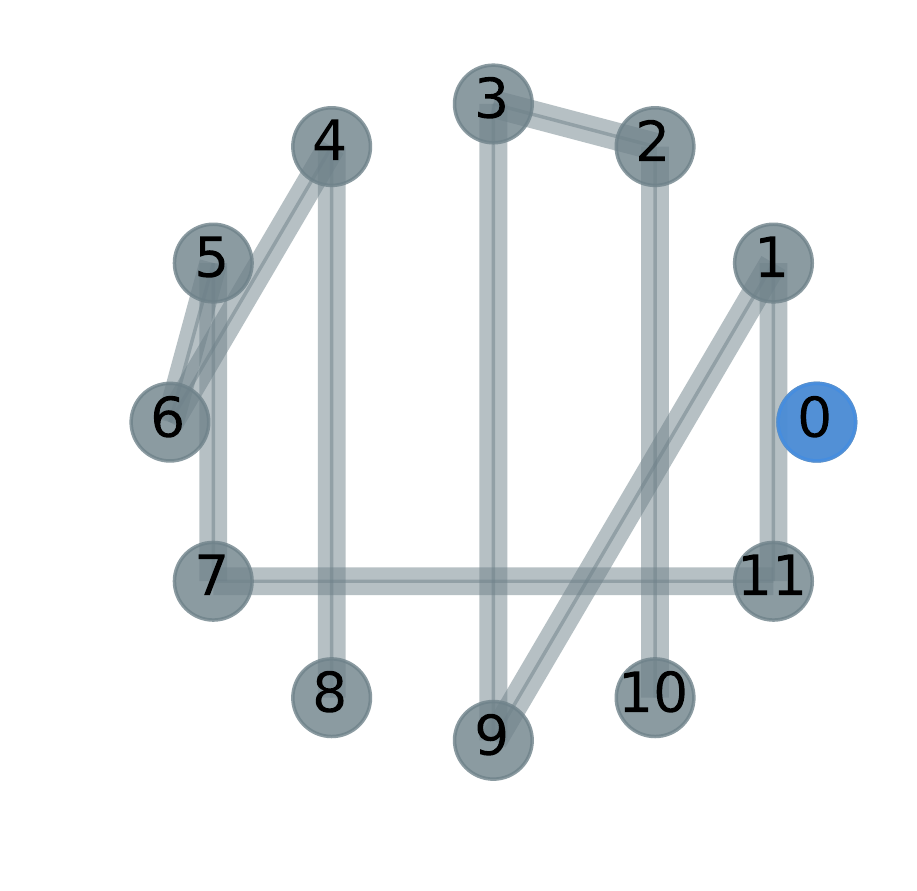}} \hfill 
\caption{Example graphs with classification results for each graph from left to right : 1, 2 and 1. This indicates the number of connected components recognized by a neural network on these graphs.}
\label{fig:graph}
\end{figure}

We constructed the learning models based on structure2vec~\citep{dai2016discriminative} for the task of determining the number of connected components (up to 3) of a given graph. 
The constructed models function as verification tools for the fidelity test. 
The idea is to first mask nodes in a graph assigned by our algorithm with different influence scores, and examine how much the classification result for this graph is affected. 
Similar fidelity tests have been widely adopted in previous research on studying feature influence~\citep{Ribeiro0G16,LundbergL17,Chen18LShapley}. 
In each trial, we first trained a neural network on the training set with random initialization. 
For each graph in the testing set, we computed the influence score for each node in this graph. 
Then we generated two data sets, $D_{\text{top}}$ and $D_{\text{bottom}}$ with graphs masked out the top-\emph{J} nodes and bottom-\emph{J} nodes (both the top-\emph{J} nodes and bottom-\emph{J} nodes are identified by their influence scores). 
We show in Figure~\ref{fig:graph} the examples of a clean graph and its top-\emph{1} and bottom-\emph{1} masked versions. 
We also constructed a third data set $D_{\text{rand}}$ by randomly masking $L$ nodes for all testing graphs. 
The evaluation results were obtained by comparing the classification performance of our trained models on these three data sets, and that achieved on the original testing sets.

\begin{figure}[t]
\begin{center}
\centerline{\includegraphics[width=0.7\columnwidth]{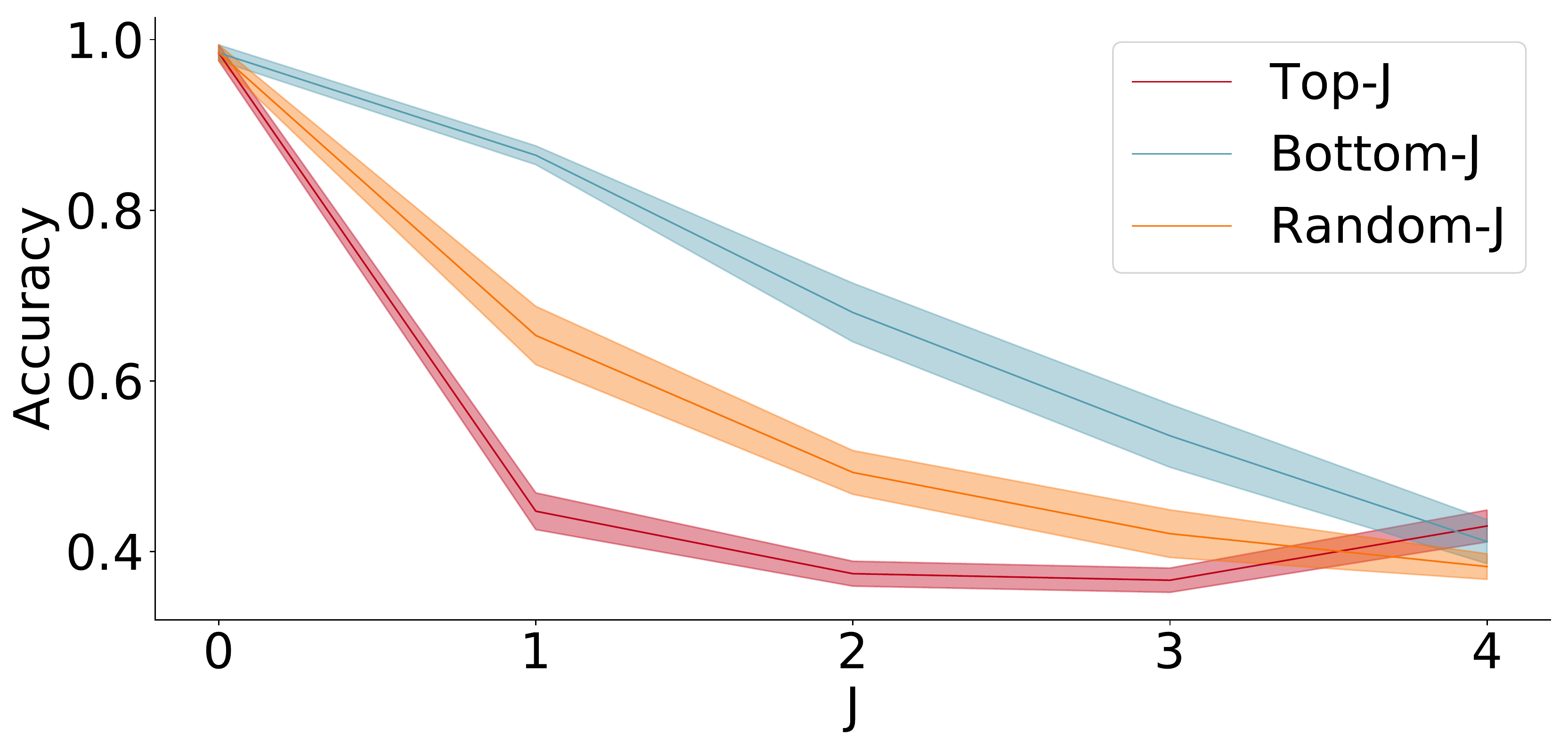}}
\caption{Accuracy of neural networks obtained on data sets with varying scales of manipulations.}
\label{fig:graph_tda}
\end{center}
\end{figure} 
We demonstrate the results in Figure~\ref{fig:graph_tda}, in which the results from all trials fit the shaded area for each line plot. 
The accuracy value indexed by $L = 0$ is the averaged accuracy of our models obtained on the clean testing sets from all trials. 
It is clear from Figure~\ref{fig:graph_tda} that the influence score calculated by our algorithm effectively indicates the impact of a node on determining the connectivity of the graph containing this node. 
In addition, as we increase $L$, the accuracy of our models obtained on $D_{\text{top}}$, $D_{\text{bottom}}$ and $D_{\text{rand}}$ degrades with different scales. 
In particular, the accuracy obtained on $D_{\text{top}}$ and $D_{\text{bottom}}$ shows the largest and smallest scales of degradation, respectively. 
The result for $D_{\text{top}}$ is surprising in that even on these simple synthetic graphs, the robustness of a neural network model is far from satisfactory. 
Specifically, similar to the results shown by \citet{dai2018adversarial}, by masking top-\emph{1} influential nodes in the testing graphs, the accuracy of a neural network is brought down to $40\% \sim 50\%$. 

\subsection{Grammar Recognition}
\label{sec:exp_grammar}

\begin{figure*}[t]
\centering
  \subfigure[SRN]{\label{fig:elman}\includegraphics[width=65mm]{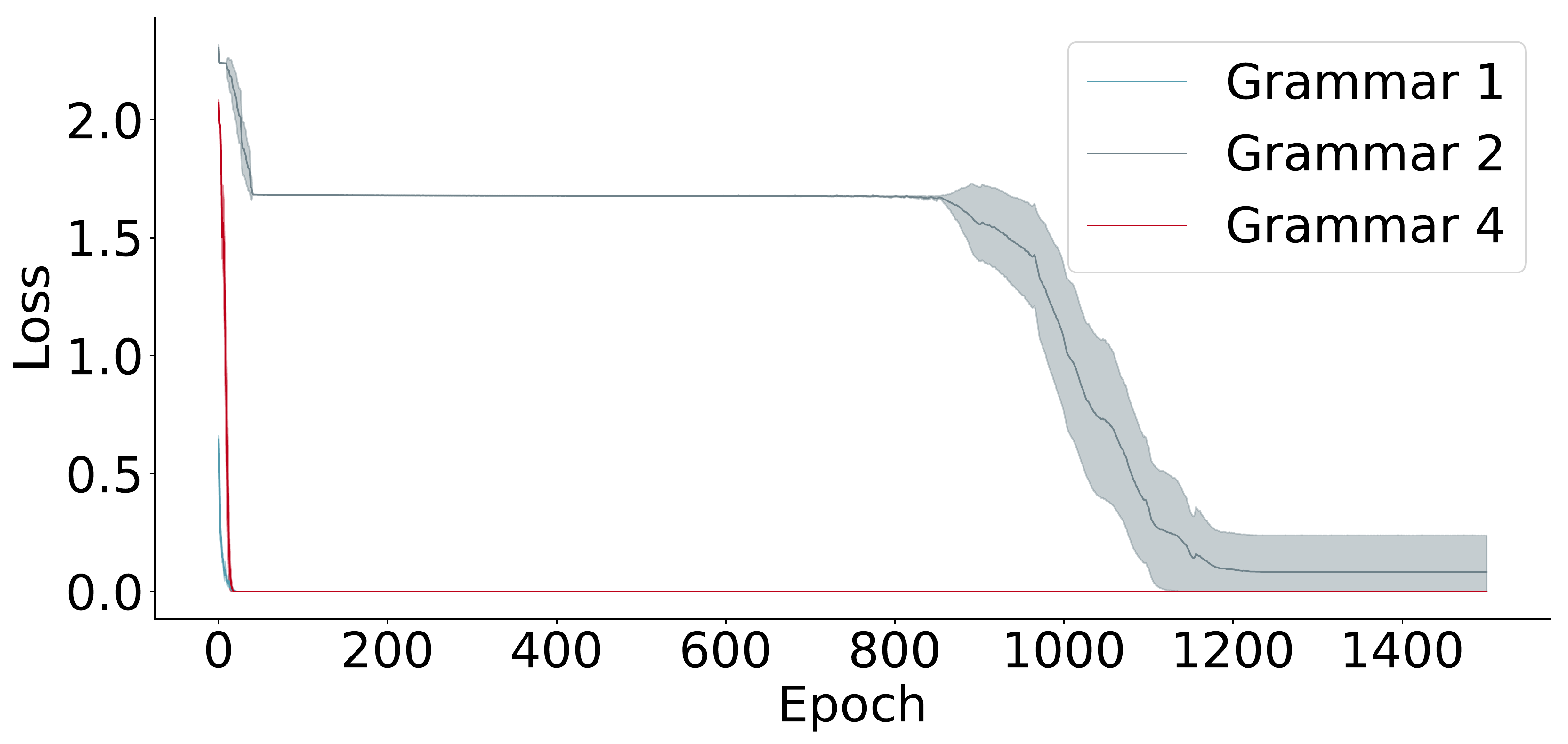}}\hfill 
  \subfigure[GRU]{\label{fig:gru}\includegraphics[width=65mm]{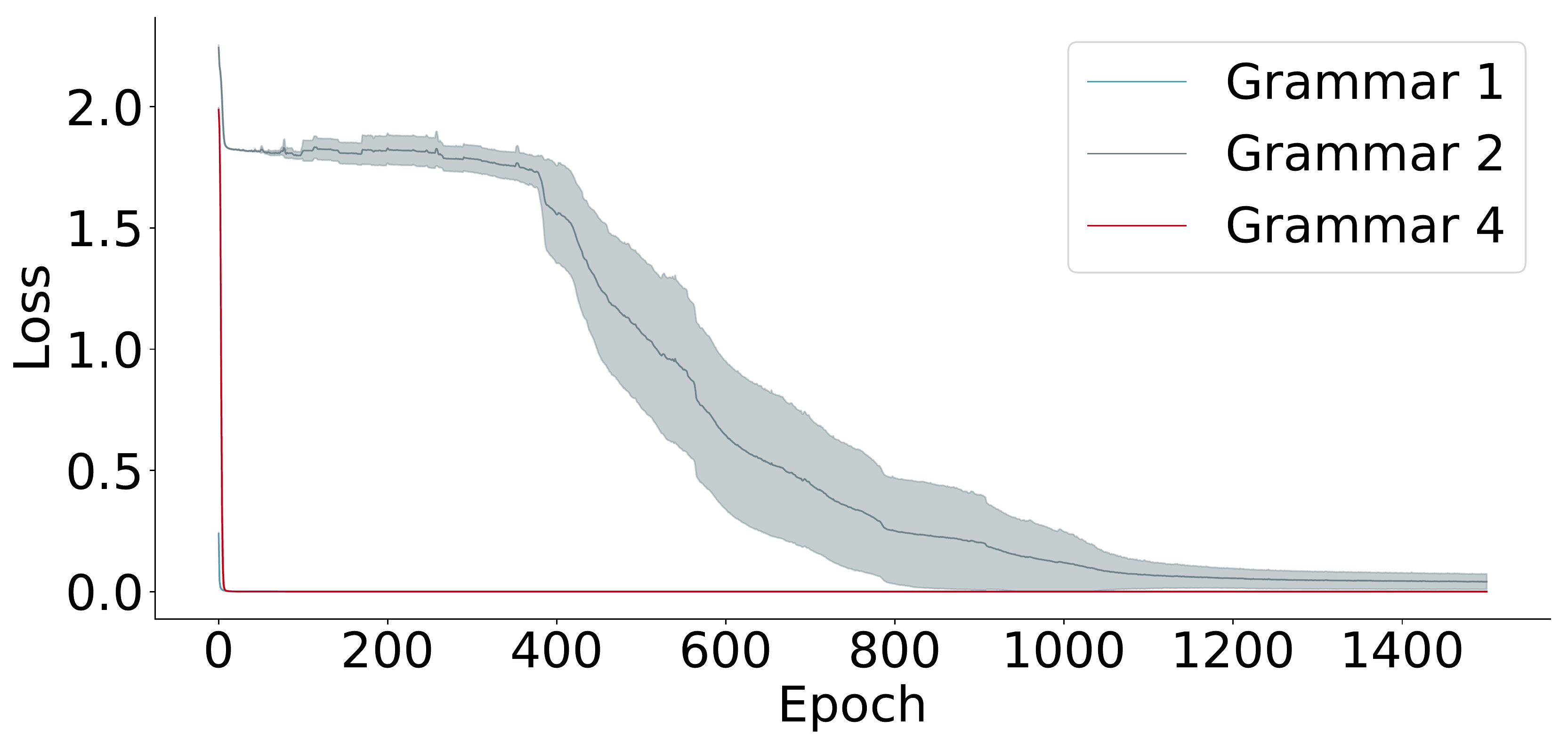}}\hfill 
  \subfigure[LSTM]{\label{fig:lstm}\includegraphics[width=65mm]{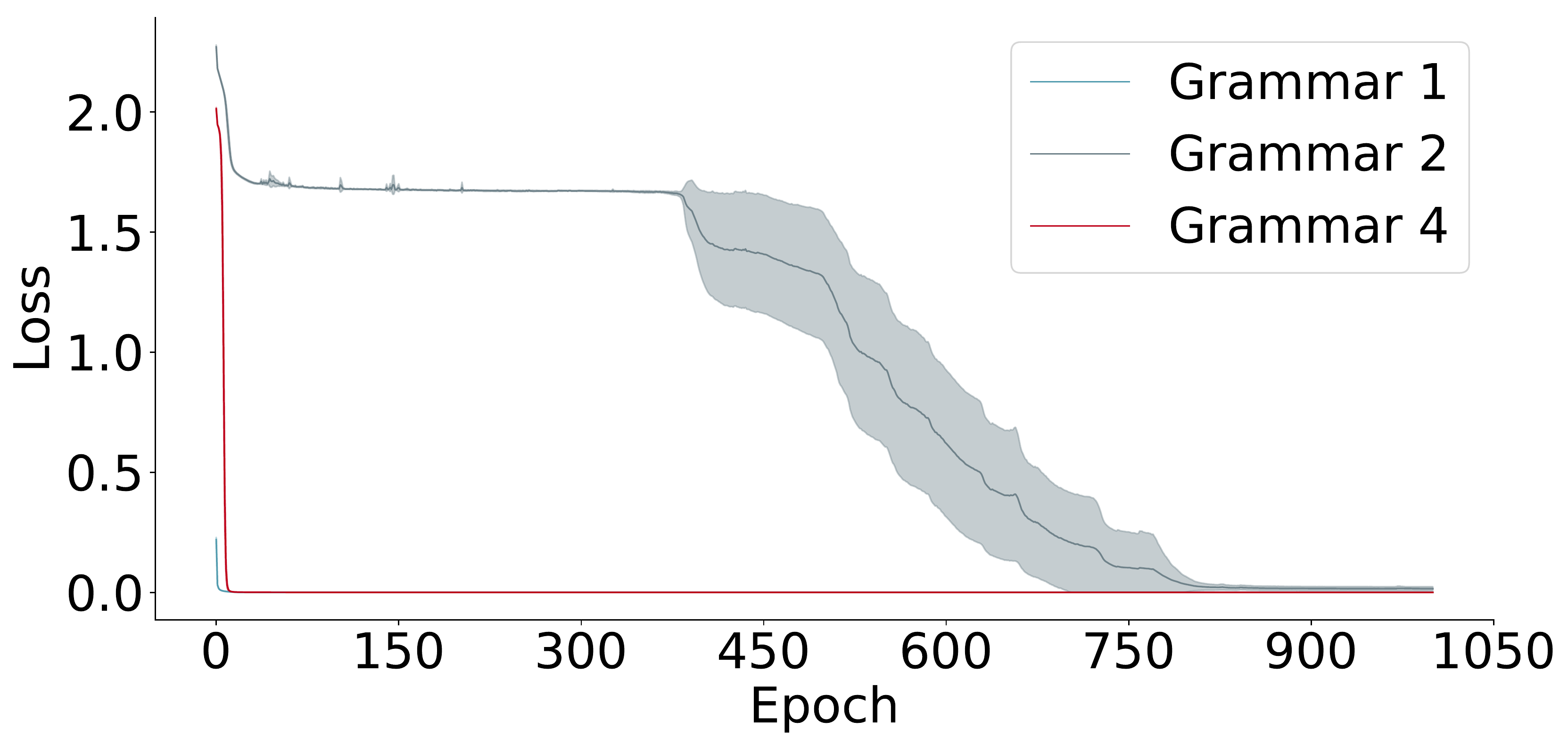}}\hfill 
  \subfigure[SRN Zoom-in]{\label{fig:elman_zoom}\includegraphics[width=65mm]{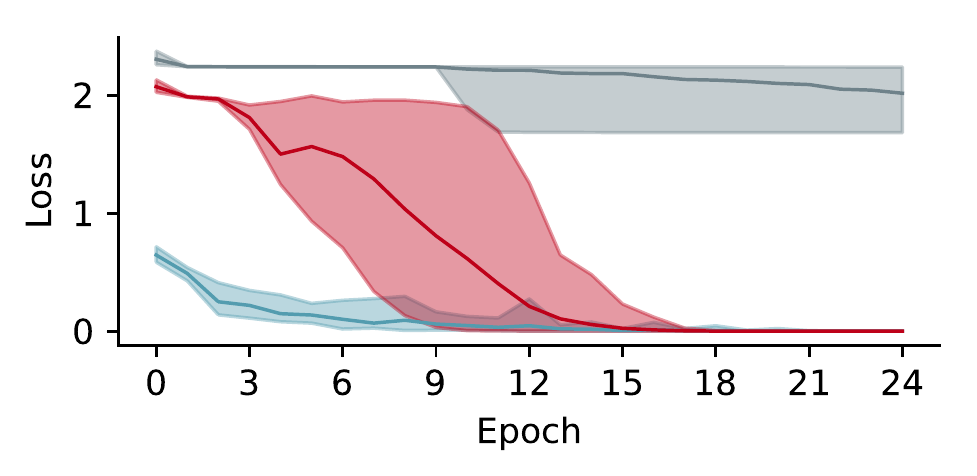}}\hfill 
  \subfigure[GRU Zoom-in]{\label{fig:gru_zoom}\includegraphics[width=65mm]{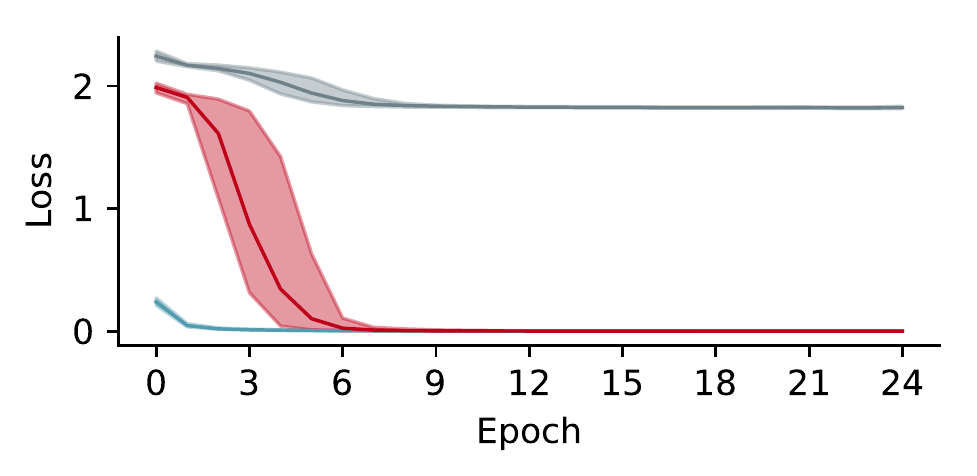}}\hfill 
  \subfigure[LSTM Zoom-in]{\label{fig:lstm_zoom}\includegraphics[width=65mm]{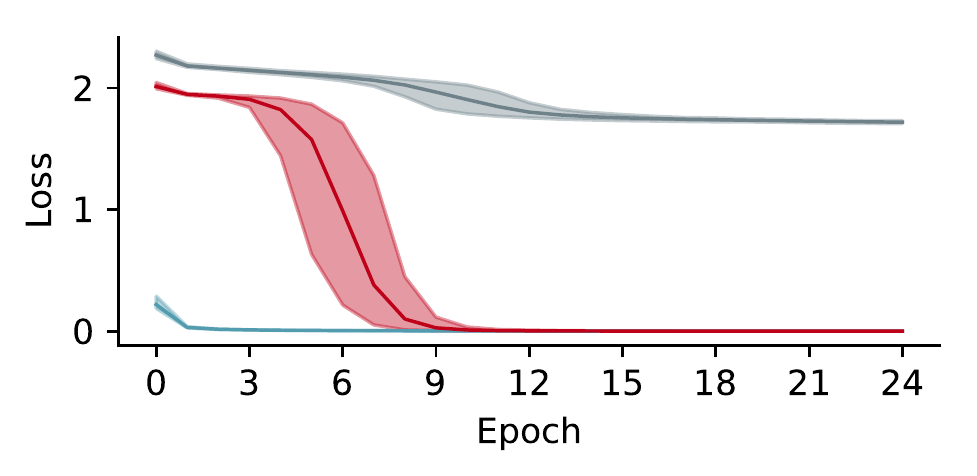}}\hfill
  
\caption{Training Performance for Different Recurrent Networks on Grammars 1, 2 and 4}
\label{fig:grammars}
\vskip 0.1in
\end{figure*}

In this set of experiments, we used three Tomita grammars introduced in Table~\ref{tab:grammars} due to their simplicity and wide adoption in various grammar learning research~\citep{weiss2017extracting,wang2018empirical,wang2018verification}. 
For each grammar, its entropy value calculated by our algorithm is shown in Table~\ref{tab:grammars}. 
We used these grammars to generate three sets of binary strings with length ranging from 2 to 13 and split the data set of each grammar into a training set and testing set with the ratio of 7 to 3. 

Then we trained several different recurrent networks (Simple Recurrent Network (SRN)~\citep{elman1990finding}, gated-recurrent-unit networks (GRU)~\citep{cho2014properties} and long-short-term-memory networks (LSTM)~\citep{hochreiter1997long}) for a binary classification task on the data set generated by each grammar. 
For each type of recurrent networks, we set the same number of parameters for models used for all three grammars to avoid the bias from these models having different learning ability. 
We trained the model of each type of RNN on each grammar for 20 trials. 
In each trial, we randomly split the training and testing set and randomly initialize the model. 
The results are demonstrated in Figure~\ref{fig:grammars}, in which the results from all trials fit the shaded area associated with each plot. 

In Figure~\ref{fig:elman},~\ref{fig:gru} and~\ref{fig:lstm}, we see that for the first and fourth grammars, which have lower entropy values, the learning process converges much faster and consistently than the process shown for the second grammar, which has the highest entropy value. This effect holds for all types of recurrent networks evaluated. To better illustrate the difference of the difficulty of learning the first and fourth grammars, we provide in Figure~\ref{fig:elman_zoom},~\ref{fig:gru_zoom} and~\ref{fig:lstm_zoom} the zoomed view for each plot at the top row of Figure~\ref{fig:grammars}. While the learning process of all models converges within ten epochs for both grammars, it is still clear that the learning process is slower for the fourth grammar. These results agree with both our analysis on the entropy of these grammars and the intuition. Specifically, the second grammar defines two sets of strings with equal cardinality when the string length is even. In this case, by flipping any binary digit of a string to its opposite (e.g., flipping a 0 to 1 or vice versa), a valid or invalid string can be converted into a string with the opposite label. This implies that a model must pay equal attention to any string to learn the underlying grammar. This corresponds to our analysis result that in the data space defined by the second grammar, each string sample shares equal influence on affecting the topological feature of this space.

\section{Conclusion}
We proposed the Shapley Homology framework to study the sample influence on topological features of a data space and its associated entropy. This provides an understanding of the intrinsic properties of both individual samples and the entire data set. We also designed an algorithm for decomposing the $0$th-Betti number using a cooperative game theory and provide analytical results for several special representative topological spaces. Furthermore, we empirically verified our results with two carefully designed experiments. We show that data points identified by our algorithm that have a larger influence on the topological features of their underlying space also have more impact on the accuracy of neural networks for determining the connectivity of their underlying graphs. We also show that a regular grammar with higher entropy has more difficulty of being learned by neural networks.


\bibliographystyle{apacite}
\bibliography{ref}

\begin{thebibliography}{}

\bibitem [\protect \citeauthoryear {%
Anirudh%
, Thiagarajan%
, Sridhar%
\BCBL {}\ \BBA {} Bremer%
}{%
Anirudh%
\ \protect \BOthers {.}}{%
{\protect \APACyear {2017}}%
}]{%
anirudh2017influential}
\APACinsertmetastar {%
anirudh2017influential}%
\begin{APACrefauthors}%
Anirudh, R.%
, Thiagarajan, J\BPBI J.%
, Sridhar, R.%
\BCBL {}\ \BBA {} Bremer, T.%
\end{APACrefauthors}%
\unskip\
\newblock
\APACrefYearMonthDay{2017}{}{}.
\newblock
{\BBOQ}\APACrefatitle {Influential Sample Selection: A Graph Signal Processing
  Approach} {Influential sample selection: A graph signal processing
  approach}.{\BBCQ}
\newblock
\APACjournalVolNumPages{arXiv preprint arXiv:1711.05407}{}{}{}.
\PrintBackRefs{\CurrentBib}

\bibitem [\protect \citeauthoryear {%
Bubenik%
}{%
Bubenik%
}{%
{\protect \APACyear {2015}}%
}]{%
bubenik2015statistical}
\APACinsertmetastar {%
bubenik2015statistical}%
\begin{APACrefauthors}%
Bubenik, P.%
\end{APACrefauthors}%
\unskip\
\newblock
\APACrefYearMonthDay{2015}{}{}.
\newblock
{\BBOQ}\APACrefatitle {Statistical topological data analysis using persistence
  landscapes} {Statistical topological data analysis using persistence
  landscapes}.{\BBCQ}
\newblock
\APACjournalVolNumPages{The Journal of Machine Learning
  Research}{16}{1}{77--102}.
\PrintBackRefs{\CurrentBib}

\bibitem [\protect \citeauthoryear {%
Carlsson%
\ \BBA {} Gabrielsson%
}{%
Carlsson%
\ \BBA {} Gabrielsson%
}{%
{\protect \APACyear {2018}}%
}]{%
Carlsson18Topological}
\APACinsertmetastar {%
Carlsson18Topological}%
\begin{APACrefauthors}%
Carlsson, G.%
\BCBT {}\ \BBA {} Gabrielsson, R\BPBI B.%
\end{APACrefauthors}%
\unskip\
\newblock
\APACrefYearMonthDay{2018}{}{}.
\newblock
{\BBOQ}\APACrefatitle {Topological Approaches to Deep Learning} {Topological
  approaches to deep learning}.{\BBCQ}
\newblock
\APACjournalVolNumPages{CoRR}{abs/1811.01122}{}{}.
\PrintBackRefs{\CurrentBib}

\bibitem [\protect \citeauthoryear {%
Carlsson%
, Ishkhanov%
, De~Silva%
\BCBL {}\ \BBA {} Zomorodian%
}{%
Carlsson%
\ \protect \BOthers {.}}{%
{\protect \APACyear {2008}}%
}]{%
carlsson2008local}
\APACinsertmetastar {%
carlsson2008local}%
\begin{APACrefauthors}%
Carlsson, G.%
, Ishkhanov, T.%
, De~Silva, V.%
\BCBL {}\ \BBA {} Zomorodian, A.%
\end{APACrefauthors}%
\unskip\
\newblock
\APACrefYearMonthDay{2008}{}{}.
\newblock
{\BBOQ}\APACrefatitle {On the local behavior of spaces of natural images} {On
  the local behavior of spaces of natural images}.{\BBCQ}
\newblock
\APACjournalVolNumPages{International journal of computer
  vision}{76}{1}{1--12}.
\PrintBackRefs{\CurrentBib}

\bibitem [\protect \citeauthoryear {%
Chazal%
\ \BBA {} Michel%
}{%
Chazal%
\ \BBA {} Michel%
}{%
{\protect \APACyear {2017}}%
}]{%
Chazal17TDA}
\APACinsertmetastar {%
Chazal17TDA}%
\begin{APACrefauthors}%
Chazal, F.%
\BCBT {}\ \BBA {} Michel, B.%
\end{APACrefauthors}%
\unskip\
\newblock
\APACrefYearMonthDay{2017}{}{}.
\newblock
{\BBOQ}\APACrefatitle {An introduction to Topological Data Analysis:
  fundamental and practical aspects for data scientists} {An introduction to
  topological data analysis: fundamental and practical aspects for data
  scientists}.{\BBCQ}
\newblock
\APACjournalVolNumPages{CoRR}{abs/1710.04019}{}{}.
\PrintBackRefs{\CurrentBib}

\bibitem [\protect \citeauthoryear {%
J.~Chen%
, Song%
, Wainwright%
\BCBL {}\ \BBA {} Jordan%
}{%
J.~Chen%
\ \protect \BOthers {.}}{%
{\protect \APACyear {2018}}%
}]{%
Chen18LShapley}
\APACinsertmetastar {%
Chen18LShapley}%
\begin{APACrefauthors}%
Chen, J.%
, Song, L.%
, Wainwright, M\BPBI J.%
\BCBL {}\ \BBA {} Jordan, M\BPBI I.%
\end{APACrefauthors}%
\unskip\
\newblock
\APACrefYearMonthDay{2018}{}{}.
\newblock
{\BBOQ}\APACrefatitle {L-Shapley and C-Shapley: Efficient Model Interpretation
  for Structured Data} {L-shapley and c-shapley: Efficient model interpretation
  for structured data}.{\BBCQ}
\newblock
\APACjournalVolNumPages{CoRR}{abs/1808.02610}{}{}.
\PrintBackRefs{\CurrentBib}

\bibitem [\protect \citeauthoryear {%
X.~Chen%
, Liu%
, Li%
, Lu%
\BCBL {}\ \BBA {} Song%
}{%
X.~Chen%
\ \protect \BOthers {.}}{%
{\protect \APACyear {2017}}%
}]{%
Chen17Backdoor}
\APACinsertmetastar {%
Chen17Backdoor}%
\begin{APACrefauthors}%
Chen, X.%
, Liu, C.%
, Li, B.%
, Lu, K.%
\BCBL {}\ \BBA {} Song, D.%
\end{APACrefauthors}%
\unskip\
\newblock
\APACrefYearMonthDay{2017}{}{}.
\newblock
{\BBOQ}\APACrefatitle {Targeted Backdoor Attacks on Deep Learning Systems Using
  Data Poisoning} {Targeted backdoor attacks on deep learning systems using
  data poisoning}.{\BBCQ}
\newblock
\APACjournalVolNumPages{CoRR}{abs/1712.05526}{}{}.
\PrintBackRefs{\CurrentBib}

\bibitem [\protect \citeauthoryear {%
Cho%
, Van~Merri{\"e}nboer%
, Bahdanau%
\BCBL {}\ \BBA {} Bengio%
}{%
Cho%
\ \protect \BOthers {.}}{%
{\protect \APACyear {2014}}%
}]{%
cho2014properties}
\APACinsertmetastar {%
cho2014properties}%
\begin{APACrefauthors}%
Cho, K.%
, Van~Merri{\"e}nboer, B.%
, Bahdanau, D.%
\BCBL {}\ \BBA {} Bengio, Y.%
\end{APACrefauthors}%
\unskip\
\newblock
\APACrefYearMonthDay{2014}{}{}.
\newblock
{\BBOQ}\APACrefatitle {On the Properties of Neural Machine Translation:
  Encoder-Decoder Approaches} {On the properties of neural machine translation:
  Encoder-decoder approaches}.{\BBCQ}
\newblock
\BIn{} \APACrefbtitle {Proceedings of SSST@EMNLP 2014, Eighth Workshop on
  Syntax, Semantics and Structure in Statistical Translation, Doha, Qatar, 25
  October 2014} {Proceedings of ssst@emnlp 2014, eighth workshop on syntax,
  semantics and structure in statistical translation, doha, qatar, 25 october
  2014}\ (\BPGS\ 103--111).
\PrintBackRefs{\CurrentBib}

\bibitem [\protect \citeauthoryear {%
Cohen-Steiner%
, Kong%
, Sohler%
\BCBL {}\ \BBA {} Valiant%
}{%
Cohen-Steiner%
\ \protect \BOthers {.}}{%
{\protect \APACyear {2018}}%
}]{%
cohen2018approximating}
\APACinsertmetastar {%
cohen2018approximating}%
\begin{APACrefauthors}%
Cohen-Steiner, D.%
, Kong, W.%
, Sohler, C.%
\BCBL {}\ \BBA {} Valiant, G.%
\end{APACrefauthors}%
\unskip\
\newblock
\APACrefYearMonthDay{2018}{}{}.
\newblock
{\BBOQ}\APACrefatitle {Approximating the Spectrum of a Graph} {Approximating
  the spectrum of a graph}.{\BBCQ}
\newblock
\BIn{} \APACrefbtitle {Proceedings of the 24th ACM SIGKDD International
  Conference on Knowledge Discovery \& Data Mining} {Proceedings of the 24th
  acm sigkdd international conference on knowledge discovery \& data mining}\
  (\BPGS\ 1263--1271).
\PrintBackRefs{\CurrentBib}

\bibitem [\protect \citeauthoryear {%
Conrad%
}{%
Conrad%
}{%
{\protect \APACyear {2008}}%
}]{%
conrad2008group}
\APACinsertmetastar {%
conrad2008group}%
\begin{APACrefauthors}%
Conrad, K.%
\end{APACrefauthors}%
\unskip\
\newblock
\APACrefYearMonthDay{2008}{}{}.
\newblock
\APACrefbtitle {Group actions.} {Group actions.}
\PrintBackRefs{\CurrentBib}

\bibitem [\protect \citeauthoryear {%
Dai%
, Dai%
\BCBL {}\ \BBA {} Song%
}{%
Dai%
\ \protect \BOthers {.}}{%
{\protect \APACyear {2016}}%
}]{%
dai2016discriminative}
\APACinsertmetastar {%
dai2016discriminative}%
\begin{APACrefauthors}%
Dai, H.%
, Dai, B.%
\BCBL {}\ \BBA {} Song, L.%
\end{APACrefauthors}%
\unskip\
\newblock
\APACrefYearMonthDay{2016}{}{}.
\newblock
{\BBOQ}\APACrefatitle {Discriminative Embeddings of Latent Variable Models for
  Structured Data} {Discriminative embeddings of latent variable models for
  structured data}.{\BBCQ}
\newblock
\APACjournalVolNumPages{arXiv preprint arXiv:1603.05629}{}{}{}.
\PrintBackRefs{\CurrentBib}

\bibitem [\protect \citeauthoryear {%
Dai%
\ \protect \BOthers {.}}{%
Dai%
\ \protect \BOthers {.}}{%
{\protect \APACyear {2018}}%
}]{%
dai2018adversarial}
\APACinsertmetastar {%
dai2018adversarial}%
\begin{APACrefauthors}%
Dai, H.%
, Li, H.%
, Tian, T.%
, Huang, X.%
, Wang, L.%
, Zhu, J.%
\BCBL {}\ \BBA {} Song, L.%
\end{APACrefauthors}%
\unskip\
\newblock
\APACrefYearMonthDay{2018}{}{}.
\newblock
{\BBOQ}\APACrefatitle {Adversarial Attack on Graph Structured Data}
  {Adversarial attack on graph structured data}.{\BBCQ}
\newblock
\APACjournalVolNumPages{}{}{}{1123--1132}.
\PrintBackRefs{\CurrentBib}

\bibitem [\protect \citeauthoryear {%
Datta%
, Sen%
\BCBL {}\ \BBA {} Zick%
}{%
Datta%
\ \protect \BOthers {.}}{%
{\protect \APACyear {2016}}%
}]{%
DattaSZ16}
\APACinsertmetastar {%
DattaSZ16}%
\begin{APACrefauthors}%
Datta, A.%
, Sen, S.%
\BCBL {}\ \BBA {} Zick, Y.%
\end{APACrefauthors}%
\unskip\
\newblock
\APACrefYearMonthDay{2016}{}{}.
\newblock
{\BBOQ}\APACrefatitle {Algorithmic Transparency via Quantitative Input
  Influence: Theory and Experiments with Learning Systems} {Algorithmic
  transparency via quantitative input influence: Theory and experiments with
  learning systems}.{\BBCQ}
\newblock
\BIn{} \APACrefbtitle {{IEEE} Symposium on Security and Privacy, {SP} 2016, San
  Jose, CA, USA, May 22-26, 2016} {{IEEE} symposium on security and privacy,
  {SP} 2016, san jose, ca, usa, may 22-26, 2016}\ (\BPGS\ 598--617).
\PrintBackRefs{\CurrentBib}

\bibitem [\protect \citeauthoryear {%
De~la Higuera%
}{%
De~la Higuera%
}{%
{\protect \APACyear {2010}}%
}]{%
de2010grammatical}
\APACinsertmetastar {%
de2010grammatical}%
\begin{APACrefauthors}%
De~la Higuera, C.%
\end{APACrefauthors}%
\unskip\
\newblock
\APACrefYear{2010}.
\newblock
\APACrefbtitle {Grammatical inference: learning automata and grammars}
  {Grammatical inference: learning automata and grammars}.
\newblock
\APACaddressPublisher{}{Cambridge University Press}.
\PrintBackRefs{\CurrentBib}

\bibitem [\protect \citeauthoryear {%
Edelsbrunner%
\ \BBA {} Harer%
}{%
Edelsbrunner%
\ \BBA {} Harer%
}{%
{\protect \APACyear {2008}}%
}]{%
edelsbrunner2008persistent}
\APACinsertmetastar {%
edelsbrunner2008persistent}%
\begin{APACrefauthors}%
Edelsbrunner, H.%
\BCBT {}\ \BBA {} Harer, J.%
\end{APACrefauthors}%
\unskip\
\newblock
\APACrefYearMonthDay{2008}{}{}.
\newblock
{\BBOQ}\APACrefatitle {Persistent homology-a survey} {Persistent homology-a
  survey}.{\BBCQ}
\newblock
\APACjournalVolNumPages{Contemporary mathematics}{453}{}{257--282}.
\PrintBackRefs{\CurrentBib}

\bibitem [\protect \citeauthoryear {%
Elman%
}{%
Elman%
}{%
{\protect \APACyear {1990}}%
}]{%
elman1990finding}
\APACinsertmetastar {%
elman1990finding}%
\begin{APACrefauthors}%
Elman, J\BPBI L.%
\end{APACrefauthors}%
\unskip\
\newblock
\APACrefYearMonthDay{1990}{}{}.
\newblock
{\BBOQ}\APACrefatitle {Finding structure in time} {Finding structure in
  time}.{\BBCQ}
\newblock
\APACjournalVolNumPages{Cognitive science}{14}{2}{179--211}.
\PrintBackRefs{\CurrentBib}

\bibitem [\protect \citeauthoryear {%
Gunning%
}{%
Gunning%
}{%
{\protect \APACyear {2017}}%
}]{%
gunning2017explainable}
\APACinsertmetastar {%
gunning2017explainable}%
\begin{APACrefauthors}%
Gunning, D.%
\end{APACrefauthors}%
\unskip\
\newblock
\APACrefYearMonthDay{2017}{}{}.
\newblock
{\BBOQ}\APACrefatitle {Explainable artificial intelligence (xai)} {Explainable
  artificial intelligence (xai)}.{\BBCQ}
\newblock
\APACjournalVolNumPages{Defense Advanced Research Projects Agency (DARPA), nd
  Web}{}{}{}.
\PrintBackRefs{\CurrentBib}

\bibitem [\protect \citeauthoryear {%
Hanneke%
}{%
Hanneke%
}{%
{\protect \APACyear {2016}}%
}]{%
hanneke2016optimal}
\APACinsertmetastar {%
hanneke2016optimal}%
\begin{APACrefauthors}%
Hanneke, S.%
\end{APACrefauthors}%
\unskip\
\newblock
\APACrefYearMonthDay{2016}{}{}.
\newblock
{\BBOQ}\APACrefatitle {The optimal sample complexity of PAC learning} {The
  optimal sample complexity of pac learning}.{\BBCQ}
\newblock
\APACjournalVolNumPages{The Journal of Machine Learning
  Research}{17}{1}{1319--1333}.
\PrintBackRefs{\CurrentBib}

\bibitem [\protect \citeauthoryear {%
Hatcher%
}{%
Hatcher%
}{%
{\protect \APACyear {2005}}%
}]{%
hatcher2005algebraic}
\APACinsertmetastar {%
hatcher2005algebraic}%
\begin{APACrefauthors}%
Hatcher, A.%
\end{APACrefauthors}%
\unskip\
\newblock
\APACrefYear{2005}.
\newblock
\APACrefbtitle {Algebraic topology} {Algebraic topology}.
\newblock
\APACaddressPublisher{}{Cambridge University Press}.
\PrintBackRefs{\CurrentBib}

\bibitem [\protect \citeauthoryear {%
Hochreiter%
\ \BBA {} Schmidhuber%
}{%
Hochreiter%
\ \BBA {} Schmidhuber%
}{%
{\protect \APACyear {1997}}%
}]{%
hochreiter1997long}
\APACinsertmetastar {%
hochreiter1997long}%
\begin{APACrefauthors}%
Hochreiter, S.%
\BCBT {}\ \BBA {} Schmidhuber, J.%
\end{APACrefauthors}%
\unskip\
\newblock
\APACrefYearMonthDay{1997}{}{}.
\newblock
{\BBOQ}\APACrefatitle {Long short-term memory} {Long short-term memory}.{\BBCQ}
\newblock
\APACjournalVolNumPages{Neural computation}{9}{8}{1735--1780}.
\PrintBackRefs{\CurrentBib}

\bibitem [\protect \citeauthoryear {%
Hofer%
, Kwitt%
, Niethammer%
\BCBL {}\ \BBA {} Uhl%
}{%
Hofer%
\ \protect \BOthers {.}}{%
{\protect \APACyear {2017}}%
}]{%
HoferKNU17}
\APACinsertmetastar {%
HoferKNU17}%
\begin{APACrefauthors}%
Hofer, C.%
, Kwitt, R.%
, Niethammer, M.%
\BCBL {}\ \BBA {} Uhl, A.%
\end{APACrefauthors}%
\unskip\
\newblock
\APACrefYearMonthDay{2017}{}{}.
\newblock
{\BBOQ}\APACrefatitle {Deep Learning with Topological Signatures} {Deep
  learning with topological signatures}.{\BBCQ}
\newblock
\BIn{} \APACrefbtitle {Advances in Neural Information Processing Systems 30:
  Annual Conference on Neural Information Processing Systems 2017, 4-9 December
  2017, Long Beach, CA, {USA}} {Advances in neural information processing
  systems 30: Annual conference on neural information processing systems 2017,
  4-9 december 2017, long beach, ca, {USA}}\ (\BPGS\ 1633--1643).
\PrintBackRefs{\CurrentBib}

\bibitem [\protect \citeauthoryear {%
Ishigami%
\ \BBA {} Tani%
}{%
Ishigami%
\ \BBA {} Tani%
}{%
{\protect \APACyear {1997}}%
}]{%
ishigami1997vc}
\APACinsertmetastar {%
ishigami1997vc}%
\begin{APACrefauthors}%
Ishigami, Y.%
\BCBT {}\ \BBA {} Tani, S.%
\end{APACrefauthors}%
\unskip\
\newblock
\APACrefYearMonthDay{1997}{}{}.
\newblock
{\BBOQ}\APACrefatitle {VC-dimensions of finite automata and commutative finite
  automata with k letters and n states} {Vc-dimensions of finite automata and
  commutative finite automata with k letters and n states}.{\BBCQ}
\newblock
\APACjournalVolNumPages{Discrete Applied Mathematics}{74}{2}{123--134}.
\PrintBackRefs{\CurrentBib}

\bibitem [\protect \citeauthoryear {%
Koh%
\ \BBA {} Liang%
}{%
Koh%
\ \BBA {} Liang%
}{%
{\protect \APACyear {2017}}%
}]{%
KohL17Influence}
\APACinsertmetastar {%
KohL17Influence}%
\begin{APACrefauthors}%
Koh, P\BPBI W.%
\BCBT {}\ \BBA {} Liang, P.%
\end{APACrefauthors}%
\unskip\
\newblock
\APACrefYearMonthDay{2017}{}{}.
\newblock
{\BBOQ}\APACrefatitle {Understanding Black-box Predictions via Influence
  Functions} {Understanding black-box predictions via influence
  functions}.{\BBCQ}
\newblock
\BIn{} \APACrefbtitle {Proceedings of the 34th International Conference on
  Machine Learning, {ICML} 2017, Sydney, NSW, Australia, 6-11 August 2017}
  {Proceedings of the 34th international conference on machine learning, {ICML}
  2017, sydney, nsw, australia, 6-11 august 2017}\ (\BPGS\ 1885--1894).
\PrintBackRefs{\CurrentBib}

\bibitem [\protect \citeauthoryear {%
Lee%
\ \BBA {} Yoo%
}{%
Lee%
\ \BBA {} Yoo%
}{%
{\protect \APACyear {2019}}%
}]{%
lee2019learning}
\APACinsertmetastar {%
lee2019learning}%
\begin{APACrefauthors}%
Lee, D.%
\BCBT {}\ \BBA {} Yoo, C\BPBI D.%
\end{APACrefauthors}%
\unskip\
\newblock
\APACrefYearMonthDay{2019}{}{}.
\newblock
\APACrefbtitle {Learning to Augment Influential Data.} {Learning to augment
  influential data.}
\newblock
\begin{APACrefURL} \url{https://openreview.net/forum?id=BygIV2CcKm}
  \end{APACrefURL}
\PrintBackRefs{\CurrentBib}

\bibitem [\protect \citeauthoryear {%
Li%
, Ovsjanikov%
\BCBL {}\ \BBA {} Chazal%
}{%
Li%
\ \protect \BOthers {.}}{%
{\protect \APACyear {2014}}%
}]{%
li2014persistence}
\APACinsertmetastar {%
li2014persistence}%
\begin{APACrefauthors}%
Li, C.%
, Ovsjanikov, M.%
\BCBL {}\ \BBA {} Chazal, F.%
\end{APACrefauthors}%
\unskip\
\newblock
\APACrefYearMonthDay{2014}{}{}.
\newblock
{\BBOQ}\APACrefatitle {Persistence-based structural recognition}
  {Persistence-based structural recognition}.{\BBCQ}
\newblock
\BIn{} \APACrefbtitle {Proceedings of the IEEE Conference on Computer Vision
  and Pattern Recognition} {Proceedings of the ieee conference on computer
  vision and pattern recognition}\ (\BPGS\ 1995--2002).
\PrintBackRefs{\CurrentBib}

\bibitem [\protect \citeauthoryear {%
Lundberg%
\ \BBA {} Lee%
}{%
Lundberg%
\ \BBA {} Lee%
}{%
{\protect \APACyear {2017}}%
}]{%
LundbergL17}
\APACinsertmetastar {%
LundbergL17}%
\begin{APACrefauthors}%
Lundberg, S\BPBI M.%
\BCBT {}\ \BBA {} Lee, S.%
\end{APACrefauthors}%
\unskip\
\newblock
\APACrefYearMonthDay{2017}{}{}.
\newblock
{\BBOQ}\APACrefatitle {A Unified Approach to Interpreting Model Predictions} {A
  unified approach to interpreting model predictions}.{\BBCQ}
\newblock
\BIn{} \APACrefbtitle {Advances in Neural Information Processing Systems 30:
  Annual Conference on Neural Information Processing Systems 2017, 4-9 December
  2017, Long Beach, CA, {USA}} {Advances in neural information processing
  systems 30: Annual conference on neural information processing systems 2017,
  4-9 december 2017, long beach, ca, {USA}}\ (\BPGS\ 4768--4777).
\PrintBackRefs{\CurrentBib}

\bibitem [\protect \citeauthoryear {%
Marsden%
}{%
Marsden%
}{%
{\protect \APACyear {2013}}%
}]{%
marsden2013eigenvalues}
\APACinsertmetastar {%
marsden2013eigenvalues}%
\begin{APACrefauthors}%
Marsden, A.%
\end{APACrefauthors}%
\unskip\
\newblock
\APACrefYearMonthDay{2013}{}{}.
\newblock
{\BBOQ}\APACrefatitle {Eigenvalues of the laplacian and their relationship to
  the connectedness of a graph} {Eigenvalues of the laplacian and their
  relationship to the connectedness of a graph}.{\BBCQ}
\newblock
\APACjournalVolNumPages{University of Chicago, REU}{}{}{}.
\PrintBackRefs{\CurrentBib}

\bibitem [\protect \citeauthoryear {%
Narahari%
}{%
Narahari%
}{%
{\protect \APACyear {2012}}%
}]{%
narahari2012shapley}
\APACinsertmetastar {%
narahari2012shapley}%
\begin{APACrefauthors}%
Narahari, Y.%
\end{APACrefauthors}%
\unskip\
\newblock
\APACrefYearMonthDay{2012}{}{}.
\newblock
{\BBOQ}\APACrefatitle {. The Shapley Value} {. the shapley value}.{\BBCQ}
\newblock

\PrintBackRefs{\CurrentBib}

\bibitem [\protect \citeauthoryear {%
Ren%
, Zeng%
, Yang%
\BCBL {}\ \BBA {} Urtasun%
}{%
Ren%
\ \protect \BOthers {.}}{%
{\protect \APACyear {2018}}%
}]{%
RenZYU18Reweight}
\APACinsertmetastar {%
RenZYU18Reweight}%
\begin{APACrefauthors}%
Ren, M.%
, Zeng, W.%
, Yang, B.%
\BCBL {}\ \BBA {} Urtasun, R.%
\end{APACrefauthors}%
\unskip\
\newblock
\APACrefYearMonthDay{2018}{}{}.
\newblock
{\BBOQ}\APACrefatitle {Learning to Reweight Examples for Robust Deep Learning}
  {Learning to reweight examples for robust deep learning}.{\BBCQ}
\newblock
\BIn{} \APACrefbtitle {Proceedings of the 35th International Conference on
  Machine Learning, {ICML} 2018, Stockholmsm{\"{a}}ssan, Stockholm, Sweden,
  July 10-15, 2018} {Proceedings of the 35th international conference on
  machine learning, {ICML} 2018, stockholmsm{\"{a}}ssan, stockholm, sweden,
  july 10-15, 2018}\ (\BPGS\ 4331--4340).
\PrintBackRefs{\CurrentBib}

\bibitem [\protect \citeauthoryear {%
Rezaei%
}{%
Rezaei%
}{%
{\protect \APACyear {2013}}%
}]{%
rezaei2013entropy}
\APACinsertmetastar {%
rezaei2013entropy}%
\begin{APACrefauthors}%
Rezaei, S\BPBI S\BPBI C.%
\end{APACrefauthors}%
\unskip\
\newblock
\APACrefYearMonthDay{2013}{}{}.
\newblock
{\BBOQ}\APACrefatitle {Entropy and graphs} {Entropy and graphs}.{\BBCQ}
\newblock
\APACjournalVolNumPages{arXiv preprint arXiv:1311.5632}{}{}{}.
\PrintBackRefs{\CurrentBib}

\bibitem [\protect \citeauthoryear {%
Ribeiro%
, Singh%
\BCBL {}\ \BBA {} Guestrin%
}{%
Ribeiro%
\ \protect \BOthers {.}}{%
{\protect \APACyear {2016}}%
}]{%
Ribeiro0G16}
\APACinsertmetastar {%
Ribeiro0G16}%
\begin{APACrefauthors}%
Ribeiro, M\BPBI T.%
, Singh, S.%
\BCBL {}\ \BBA {} Guestrin, C.%
\end{APACrefauthors}%
\unskip\
\newblock
\APACrefYearMonthDay{2016}{}{}.
\newblock
{\BBOQ}\APACrefatitle {"Why Should {I} Trust You?": Explaining the Predictions
  of Any Classifier} {"why should {I} trust you?": Explaining the predictions
  of any classifier}.{\BBCQ}
\newblock
\BIn{} \APACrefbtitle {Proceedings of the 22nd {ACM} {SIGKDD} International
  Conference on Knowledge Discovery and Data Mining, San Francisco, CA, USA,
  August 13-17, 2016} {Proceedings of the 22nd {ACM} {SIGKDD} international
  conference on knowledge discovery and data mining, san francisco, ca, usa,
  august 13-17, 2016}\ (\BPGS\ 1135--1144).
\PrintBackRefs{\CurrentBib}

\bibitem [\protect \citeauthoryear {%
Rote%
\ \BBA {} Vegter%
}{%
Rote%
\ \BBA {} Vegter%
}{%
{\protect \APACyear {2006}}%
}]{%
rote2006computational}
\APACinsertmetastar {%
rote2006computational}%
\begin{APACrefauthors}%
Rote, G.%
\BCBT {}\ \BBA {} Vegter, G.%
\end{APACrefauthors}%
\unskip\
\newblock
\APACrefYearMonthDay{2006}{}{}.
\newblock
{\BBOQ}\APACrefatitle {Computational topology: An introduction} {Computational
  topology: An introduction}.{\BBCQ}
\newblock
\BIn{} \APACrefbtitle {Effective Computational Geometry for curves and
  surfaces} {Effective computational geometry for curves and surfaces}\ (\BPGS\
  277--312).
\newblock
\APACaddressPublisher{}{Springer}.
\PrintBackRefs{\CurrentBib}

\bibitem [\protect \citeauthoryear {%
Schapira%
}{%
Schapira%
}{%
{\protect \APACyear {2001}}%
}]{%
schapira2001categories}
\APACinsertmetastar {%
schapira2001categories}%
\begin{APACrefauthors}%
Schapira, P.%
\end{APACrefauthors}%
\unskip\
\newblock
\APACrefYear{2001}.
\newblock
\APACrefbtitle {Categories and homological algebra} {Categories and homological
  algebra}.
\newblock
\APACaddressPublisher{}{Soci{\'e}t{\'e} math{\'e}tique de France}.
\PrintBackRefs{\CurrentBib}

\bibitem [\protect \citeauthoryear {%
Tomita%
}{%
Tomita%
}{%
{\protect \APACyear {1982}}%
}]{%
tomita1982dynamic}
\APACinsertmetastar {%
tomita1982dynamic}%
\begin{APACrefauthors}%
Tomita, M.%
\end{APACrefauthors}%
\unskip\
\newblock
\APACrefYearMonthDay{1982}{}{}.
\newblock
{\BBOQ}\APACrefatitle {Dynamic construction of finite-state automata from
  examples using hill-climbing.} {Dynamic construction of finite-state automata
  from examples using hill-climbing.}{\BBCQ}
\newblock
\BIn{} \APACrefbtitle {Proceedings of the Fourth Annual Conference of the
  Cognitive Science Society} {Proceedings of the fourth annual conference of
  the cognitive science society}\ (\BPGS\ 105--108).
\PrintBackRefs{\CurrentBib}

\bibitem [\protect \citeauthoryear {%
Turner%
, Mukherjee%
\BCBL {}\ \BBA {} Boyer%
}{%
Turner%
\ \protect \BOthers {.}}{%
{\protect \APACyear {2014}}%
}]{%
turner2014persistent}
\APACinsertmetastar {%
turner2014persistent}%
\begin{APACrefauthors}%
Turner, K.%
, Mukherjee, S.%
\BCBL {}\ \BBA {} Boyer, D\BPBI M.%
\end{APACrefauthors}%
\unskip\
\newblock
\APACrefYearMonthDay{2014}{}{}.
\newblock
{\BBOQ}\APACrefatitle {Persistent homology transform for modeling shapes and
  surfaces} {Persistent homology transform for modeling shapes and
  surfaces}.{\BBCQ}
\newblock
\APACjournalVolNumPages{Information and Inference: A Journal of the
  IMA}{3}{4}{310--344}.
\PrintBackRefs{\CurrentBib}

\bibitem [\protect \citeauthoryear {%
Vapnik%
}{%
Vapnik%
}{%
{\protect \APACyear {2000}}%
}]{%
VCdimension}
\APACinsertmetastar {%
VCdimension}%
\begin{APACrefauthors}%
Vapnik, V\BPBI N.%
\end{APACrefauthors}%
\unskip\
\newblock
\APACrefYear{2000}.
\newblock
\APACrefbtitle {The Nature of Statistical Learning Theory, Second Edition} {The
  nature of statistical learning theory, second edition}.
\newblock
\APACaddressPublisher{}{Springer}.
\PrintBackRefs{\CurrentBib}

\bibitem [\protect \citeauthoryear {%
Q.~Wang%
, Zhang%
, Liu%
\BCBL {}\ \BBA {} Giles%
}{%
Q.~Wang%
, Zhang%
, Liu%
\BCBL {}\ \BBA {} Giles%
}{%
{\protect \APACyear {2018}}%
}]{%
wang2018verification}
\APACinsertmetastar {%
wang2018verification}%
\begin{APACrefauthors}%
Wang, Q.%
, Zhang, K.%
, Liu, X.%
\BCBL {}\ \BBA {} Giles, C\BPBI L.%
\end{APACrefauthors}%
\unskip\
\newblock
\APACrefYearMonthDay{2018}{}{}.
\newblock
{\BBOQ}\APACrefatitle {Verification of Recurrent Neural Networks Through Rule
  Extraction} {Verification of recurrent neural networks through rule
  extraction}.{\BBCQ}
\newblock
\APACjournalVolNumPages{arXiv preprint arXiv:1811.06029}{}{}{}.
\PrintBackRefs{\CurrentBib}

\bibitem [\protect \citeauthoryear {%
Q.~Wang%
, Zhang%
, Ororbia%
\BCBL {}\ \protect \BOthers {.}}{%
Q.~Wang%
, Zhang%
, Ororbia%
\BCBL {}\ \protect \BOthers {.}}{%
{\protect \APACyear {2018}}%
}]{%
wang2018comparative}
\APACinsertmetastar {%
wang2018comparative}%
\begin{APACrefauthors}%
Wang, Q.%
, Zhang, K.%
, Ororbia, I.%
, Alexander, G.%
, Xing, X.%
, Liu, X.%
\BCBL {}\ \BBA {} Giles, C\BPBI L.%
\end{APACrefauthors}%
\unskip\
\newblock
\APACrefYearMonthDay{2018}{}{}.
\newblock
{\BBOQ}\APACrefatitle {A Comparative Study of Rule Extraction for Recurrent
  Neural Networks} {A comparative study of rule extraction for recurrent neural
  networks}.{\BBCQ}
\newblock
\APACjournalVolNumPages{arXiv preprint arXiv:1801.05420}{}{}{}.
\PrintBackRefs{\CurrentBib}

\bibitem [\protect \citeauthoryear {%
Q.~Wang%
, Zhang%
, Ororbia~II%
\BCBL {}\ \protect \BOthers {.}}{%
Q.~Wang%
, Zhang%
, Ororbia~II%
\BCBL {}\ \protect \BOthers {.}}{%
{\protect \APACyear {2018}}%
}]{%
wang2018empirical}
\APACinsertmetastar {%
wang2018empirical}%
\begin{APACrefauthors}%
Wang, Q.%
, Zhang, K.%
, Ororbia~II, A\BPBI G.%
, Xing, X.%
, Liu, X.%
\BCBL {}\ \BBA {} Giles, C\BPBI L.%
\end{APACrefauthors}%
\unskip\
\newblock
\APACrefYearMonthDay{2018}{}{}.
\newblock
{\BBOQ}\APACrefatitle {An empirical evaluation of rule extraction from
  recurrent neural networks} {An empirical evaluation of rule extraction from
  recurrent neural networks}.{\BBCQ}
\newblock
\APACjournalVolNumPages{Neural computation}{30}{9}{2568--2591}.
\PrintBackRefs{\CurrentBib}

\bibitem [\protect \citeauthoryear {%
T.~Wang%
, Zhu%
, Torralba%
\BCBL {}\ \BBA {} Efros%
}{%
T.~Wang%
\ \protect \BOthers {.}}{%
{\protect \APACyear {2018}}%
}]{%
Wang18Distillation}
\APACinsertmetastar {%
Wang18Distillation}%
\begin{APACrefauthors}%
Wang, T.%
, Zhu, J.%
, Torralba, A.%
\BCBL {}\ \BBA {} Efros, A\BPBI A.%
\end{APACrefauthors}%
\unskip\
\newblock
\APACrefYearMonthDay{2018}{}{}.
\newblock
{\BBOQ}\APACrefatitle {Dataset Distillation} {Dataset distillation}.{\BBCQ}
\newblock
\APACjournalVolNumPages{CoRR}{abs/1811.10959}{}{}.
\PrintBackRefs{\CurrentBib}

\bibitem [\protect \citeauthoryear {%
Y.~Wang%
\ \BBA {} Chaudhuri%
}{%
Y.~Wang%
\ \BBA {} Chaudhuri%
}{%
{\protect \APACyear {2018}}%
}]{%
Wang18Poisoning}
\APACinsertmetastar {%
Wang18Poisoning}%
\begin{APACrefauthors}%
Wang, Y.%
\BCBT {}\ \BBA {} Chaudhuri, K.%
\end{APACrefauthors}%
\unskip\
\newblock
\APACrefYearMonthDay{2018}{}{}.
\newblock
{\BBOQ}\APACrefatitle {Data Poisoning Attacks against Online Learning} {Data
  poisoning attacks against online learning}.{\BBCQ}
\newblock
\APACjournalVolNumPages{CoRR}{abs/1808.08994}{}{}.
\PrintBackRefs{\CurrentBib}

\bibitem [\protect \citeauthoryear {%
Weiss%
, Goldberg%
\BCBL {}\ \BBA {} Yahav%
}{%
Weiss%
\ \protect \BOthers {.}}{%
{\protect \APACyear {2017}}%
}]{%
weiss2017extracting}
\APACinsertmetastar {%
weiss2017extracting}%
\begin{APACrefauthors}%
Weiss, G.%
, Goldberg, Y.%
\BCBL {}\ \BBA {} Yahav, E.%
\end{APACrefauthors}%
\unskip\
\newblock
\APACrefYearMonthDay{2017}{}{}.
\newblock
{\BBOQ}\APACrefatitle {Extracting Automata from Recurrent Neural Networks Using
  Queries and Counterexamples} {Extracting automata from recurrent neural
  networks using queries and counterexamples}.{\BBCQ}
\newblock
\APACjournalVolNumPages{arXiv preprint arXiv:1711.09576}{}{}{}.
\PrintBackRefs{\CurrentBib}

\bibitem [\protect \citeauthoryear {%
Williams%
}{%
Williams%
}{%
{\protect \APACyear {2004}}%
}]{%
williams2004introduction}
\APACinsertmetastar {%
williams2004introduction}%
\begin{APACrefauthors}%
Williams, S\BPBI G.%
\end{APACrefauthors}%
\unskip\
\newblock
\APACrefYearMonthDay{2004}{}{}.
\newblock
{\BBOQ}\APACrefatitle {Introduction to symbolic dynamics} {Introduction to
  symbolic dynamics}.{\BBCQ}
\newblock
\BIn{} \APACrefbtitle {Proceedings of Symposia in Applied Mathematics}
  {Proceedings of symposia in applied mathematics}\ (\BVOL~60, \BPGS\ 1--12).
\PrintBackRefs{\CurrentBib}

\bibitem [\protect \citeauthoryear {%
Yeh%
, Kim%
, Yen%
\BCBL {}\ \BBA {} Ravikumar%
}{%
Yeh%
\ \protect \BOthers {.}}{%
{\protect \APACyear {2018}}%
}]{%
YehKYR18Representer}
\APACinsertmetastar {%
YehKYR18Representer}%
\begin{APACrefauthors}%
Yeh, C.%
, Kim, J\BPBI S.%
, Yen, I\BPBI E.%
\BCBL {}\ \BBA {} Ravikumar, P.%
\end{APACrefauthors}%
\unskip\
\newblock
\APACrefYearMonthDay{2018}{}{}.
\newblock
{\BBOQ}\APACrefatitle {Representer Point Selection for Explaining Deep Neural
  Networks} {Representer point selection for explaining deep neural
  networks}.{\BBCQ}
\newblock
\BIn{} \APACrefbtitle {Advances in Neural Information Processing Systems 31:
  Annual Conference on Neural Information Processing Systems 2018, NeurIPS
  2018, 3-8 December 2018, Montr{\'{e}}al, Canada.} {Advances in neural
  information processing systems 31: Annual conference on neural information
  processing systems 2018, neurips 2018, 3-8 december 2018, montr{\'{e}}al,
  canada.}\ (\BPGS\ 9311--9321).
\PrintBackRefs{\CurrentBib}

\end{thebibliography}

\newpage
\appendix
\section{Proof of the Results for the Six Special Sets of Graphs}

Here we provide proofs for results shown in Table~\ref{tab:special graph results}.\footnote{Only the calculation of Shapley value is given and influence score can be easily obtained by normalization of Shapley value.} 
Further with these results one can easily derive corollaries by some basic inequalities. 

\paragraph{The complete graph and cycles}
For the complete graph and cycles, the results presented in Table~\ref{tab:special graph results} are trivial since these two graphs preserve invariance under permutation. 

\paragraph{The wheel graph}
For wheel graph shown in Table~\ref{tab:special graph results}, we have 
\begin{equation}
\begin{aligned}
\nonumber &\textup{Periphery:} \;\;\frac{1}{N!}(\sum_{k=0}^{N-4}\binom{N-4}{k}k!(N-k-1)!+\sum_{k=0}^{N-5}\binom{N-4}{k}(k+2)!(N-k-3)!),\\  &\textup{Center:} \;\;\frac{1}{N}+\frac{1}{N!}\sum_{m=2}^{N-3}\sum_{k=2}^{l}T(N-1,k,m)(k-1)m!(N-m-1)!,
 \end{aligned}
\end{equation}
where $l=\min(m,N-m-1)$ and $T(N,k,m)=\frac{N}{m}\binom{m}{k}\binom{N-m-1}{k-1}$, and simplifying the above formulas gives the results in Table~\ref{tab:special graph results}. 

\paragraph{The star graph}
For the star graph, we have
\begin{equation}
\begin{aligned}
\nonumber &\textup{Periphery:} \;\;\frac{1}{N!}\sum_{k=0}^{N-2}\binom{N-2}{k}k!(N-k-1)!=\frac{1}{2},\\  &\textup{Center:} \;\;\frac{1}{N}+\frac{1}{N!}\sum_{k=2}^{N-1}\binom{N-1}{k}(k-1)(k-1)!(N-k)!=\frac{N^2-3N+4}{2N}.
 \end{aligned}
\end{equation}

\paragraph{The path graph} 
For path graph we have
\begin{equation}
\begin{aligned}
\nonumber &\textup{Ends:} \;\;\frac{1}{N!}\sum_{k=0}^{N-2}\binom{N-2}{k}k!(N-k-1)!=\frac{1}{2},\\  &\textup{Middle:} \;\;\frac{2}{N!}\sum_{k=0}^{N-3}\binom{N-3}{k}k!(N-k-1)!=\frac{2}{3}.
 \end{aligned}
\end{equation}

\paragraph{The complete bipartite graph} 
For complete bipartite graph we have
\begin{equation}
\begin{aligned}
\nonumber &\textup{m side:} \;\;\frac{1}{(m+n)!}(\sum_{k=0}^{m-1}\binom{m-1}{k}k!(m+n-k-1)!+\sum_{k=2}^{n}\binom{n}{k}(k-1)k!(m+n-k-1)!),\\  &\textup{n side:} \;\;\frac{1}{(m+n)!}(\sum_{k=0}^{n-1}\binom{n-1}{k}k!(m+n-k-1)!+\sum_{k=2}^{m}\binom{m}{k}(k-1)k!(m+n-k-1)!),
 \end{aligned}
\end{equation}
which gives the results in Table~\ref{tab:special graph results} by simplifying the above formulations. 

Above derivation has applied the following facts which can be easily verified by combinatorics:
\begin{equation}
\nonumber \frac{1}{N!}\sum_{k=0}^{N-m}\binom{N-m}{k}k!(N-k-1)!=\frac{1}{m},
\end{equation}
\begin{equation}
\nonumber \frac{1}{(m+n)!}\sum_{k=2}^{m}\binom{m}{k}(k-1)k!(m+n-k-1)!=\frac{m(m-1)}{n(n+1)(m+n)},
\end{equation}
\begin{equation}
\nonumber \frac{1}{N!}\sum_{m=2}^{N-3}\sum_{k=2}^{l}T(N-1,k,m)(k-1)m!(N-m-1)!=\frac{(N-3)(N-4)}{6N},
\end{equation}
where $l=\min(m,N-m-1)$ and $T(N,k,m)=\frac{N}{m}\binom{m}{k}\binom{N-m-1}{k-1}$.
\newpage

\section{Experiment and Model Settings}
Here we provide all the parameters set in the experiments.
\begin{table}[ht!]
\small
\centering
\begin{tabular}{|c|c|c|c|c|c|c|}
\hline \hline
Experiment                                                                  & \multicolumn{2}{c|}{Graph}                                                                                                         & \multicolumn{4}{c|}{Grammar}                                          \\ \hline \hline
\multirow{5}{*}{\begin{tabular}[c]{@{}c@{}}Model \\ Setting\end{tabular}}   & Component                                                                 & Hidden                                                 & Model                   & \multicolumn{3}{c|}{Hidden (Total)}         \\ \cline{2-7} 
                                                                            & \multirow{2}{*}{Struct2vec}                                               & \multirow{2}{*}{32}                                    & SRN                     & \multicolumn{3}{c|}{200 (40800)}            \\
                                                                            &                                                                           &                                                        & GRU                     & \multicolumn{3}{c|}{80 (19920)}             \\
                                                                            & \multirow{2}{*}{\begin{tabular}[c]{@{}c@{}}Neural\\ Network\end{tabular}} & \multirow{2}{*}{16}                                    & LSTM                    & \multicolumn{3}{c|}{80 (19920)}             \\ \cline{4-7} 
                                                                            &                                                                           &                                                        & Optimizer               & \multicolumn{3}{c|}{Rmsprop}                \\ \hline
\multirow{7}{*}{\begin{tabular}[c]{@{}c@{}}Dataset \\ Setting\end{tabular}} & Class                                                                     & 3                                                      & Class                   & \multicolumn{3}{c|}{2}                      \\ \cline{2-7} 
                                                                            & \multirow{2}{*}{Total}                                                    & \multirow{2}{*}{6000}                                  & \multirow{2}{*}{Total}  & G1            & G2            & G4          \\
                                                                            &                                                                           &                                                        &                         & 8677          & 8188          & 8188        \\ \cline{2-7} 
                                                                            & \begin{tabular}[c]{@{}c@{}}Training\\ (per class)\end{tabular}            & \begin{tabular}[c]{@{}c@{}}5400\\ (1800)\end{tabular}  & Training                & 6073          & 5730          & 5730        \\ \cline{2-7} 
                                                                            & \begin{tabular}[c]{@{}c@{}}Testing\\ (per class)\end{tabular}             & \begin{tabular}[c]{@{}c@{}}600\\ (200)\end{tabular}    & Testing                 & 2604          & 2458          & 2458        \\ \cline{2-7} 
                                                                            & Prob.                                                                     & \begin{tabular}[c]{@{}c@{}}0.02 -  0.21\end{tabular} & \multirow{2}{*}{Length} & \multicolumn{3}{c|}{\multirow{2}{*}{2- 13}} \\ \cline{2-3}
                                                                            & Nodes                                                                     & 8-14                                                   &                         & \multicolumn{3}{c|}{}                       \\ \hline \hline
\end{tabular}
\caption{Parameter Settings For Neural Network Models and Datasets.}
\end{table}

\end{document}